\documentclass[onecolumn,10pt]{asme2ej}
\usepackage{cite}
\usepackage{subcaption} %
\usepackage{amsmath,amssymb,amsfonts}
\usepackage{tabularx, booktabs, algorithmic, graphicx, textcomp, xcolor, multirow, mathrsfs, caption, array, mathtools, arydshln, manyfoot, algorithm, listings}
\usepackage{url}
\usepackage[normalem]{ulem}

\usepackage{bm}
\newcommand{\bs}[1]{\bm{#1}}

\title{Teleoperation of Continuum Instruments: Investigation of Linear vs. Angular Commands Through Task-Priority Analysis}

\title{Teleoperation of Continuum Instruments: Task-Priority Analysis of Linear–Angular Command Interplay}

\author{Ehsan Nasiri\\
\textit{Mechanical Engineering Department}\\
Stevens Institute of Technology, Hoboken, NJ, USA\\
enasiri@stevens.edu
\and
{\textbf{Long Wang}}\\
\textit{Mechanical Engineering Department}\\
Stevens Institute of Technology, Hoboken, NJ, USA\\
lwang4@stevens.edu
}

\begin{document}
\maketitle

\begin{abstract}
This paper addresses the challenge of teleoperating continuum instruments for minimally invasive surgery (MIS). We develop and adopt a novel task-priority-based kinematic formulation to quantitatively investigate teleoperation commands for continuum instruments under remote center of motion (RCM) constraints. Using redundancy resolution methods, we investigate the kinematic performance during teleoperation, comparing linear and angular commands within a task-priority scheme. For experimental validation, an instrument module (IM) was designed and integrated with a 7-DoF manipulator.
Assessments, simulations, and experimental validations demonstrated the effectiveness of the proposed framework. The experiments involved several tasks: trajectory tracking of the IM tip along multiple paths with varying priorities for linear and angular teleoperation commands, pushing a ball along predefined paths on a silicon board, following a pattern on a pegboard, and guiding the continuum tip through rings on a ring board using a standard surgical kit.
\end{abstract}

\section{Introduction}
Flexible surgical instruments are crucial in minimally invasive surgeries (MIS), particularly in transluminal procedures, and have seen considerable advancements recently. Compared to open surgeries, MIS offers significant benefits, including shorter recovery times, minimal trauma due to tiny incisions, and reduced risk of inflammation. Traditional tools such as laparoscopes, endoscopes, cystoscopes, and hysteroscopes are effective in MIS. However, their rigid and lengthy mechanical structures limit dexterity and sensitivity, restricting broader applications. The rigid and elongated characteristics of these instruments also prevent navigation through natural orifices like the intestinal and bronchial tracts. In contrast, the inherent flexibility of continuum surgical instruments allows for safer interactions with anatomical structures. Consequently, these instruments are emerging as promising alternatives to rigid tools in MIS applications. Unlike traditional robots, continuum robots typically feature a snake-like shape with continuous backbones. These robots possess bending, flexible structures with an infinite degree of freedom, which inherently enhances safety and improves intracorporeal dexterity\cite{Bajo2013,Bajo2013_2,dong2016novel}.

In recent years, advancements in robotic-assisted procedures have garnered significant attention from both industry and academic groups. Efforts have focused on enhancing and diversifying the applications of Teleoperational systems, particularly in single-port access surgery, natural orifice translumenal endoscopic surgery etc\cite{DiMaio2010,DiMaio2011,Ikuta2003,burgner2011}. Design variations in these systems range from purely rigid-link  to wire-actuated rigid link endeffectors and continuum instrument \cite{webster2010design,bajo2012integration}.
Effective telemanipulation and control of continuum robots present a range of complex challenges, including kinematic modeling, the real-time execution of direct and inverse kinematics, backlash compensation, friction estimation, and extending actuation lines. Numerous strategies have been developed to overcome these issues and constraints.
The mathematical foundations for modeling and analyzing hyper-redundant robots, as well as the kinematics of these robots, were established by works like \cite{chirikjian1994modal,zanganeh1995inverse} and \cite{Xu2008,Ding2013,goldman2014compliant}.
\begin{figure*}
	\centering
	\includegraphics[width=\textwidth, height= 0.6 \textwidth]{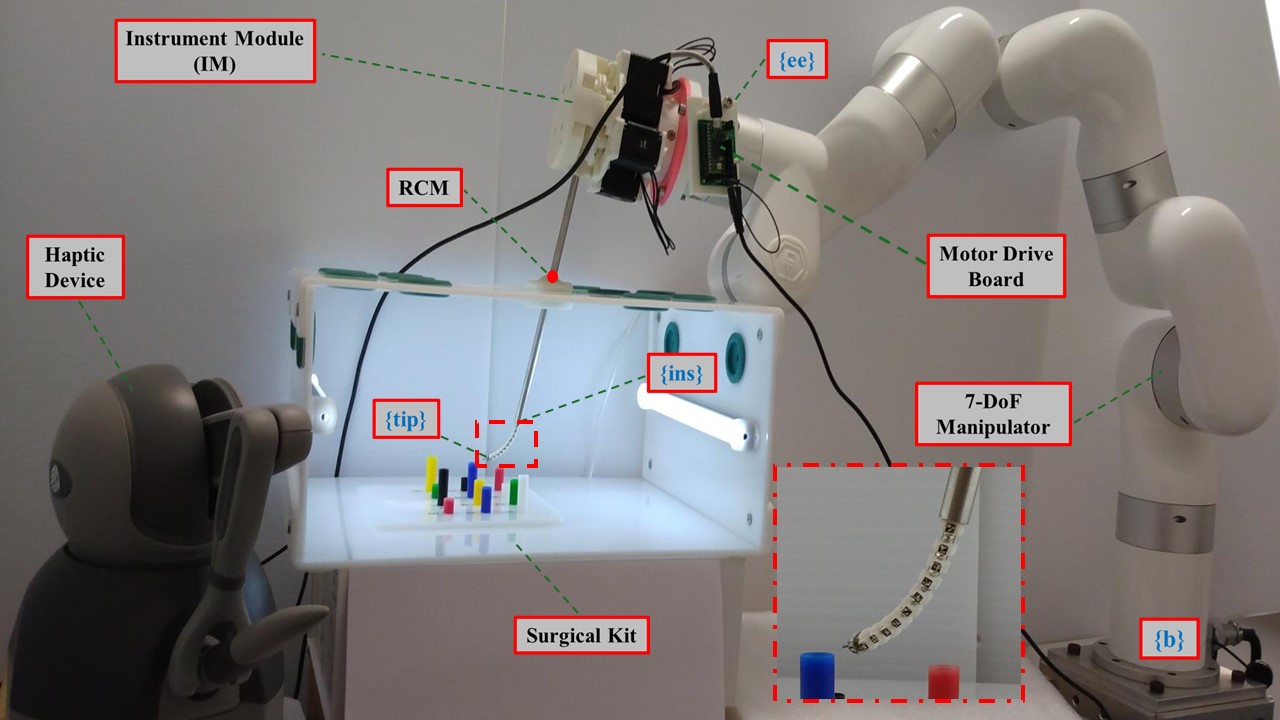}
	\caption{System architecture includes a 7-DoF manipulator, custom-designed instrument module (IM), haptic device, and the surgical kit}
	\label{fig:System_arch}
 \vspace{-0.2cm}
\end{figure*}
Since the early introduction of continuum robots in research, such as in \cite{Robinson99}, researchers have been drawn to their flexibility, dexterity, and ability to operate effectively in diverse environments. These robots can safely navigate complex and delicate settings while responding accurately to the operator's commands.
In the context of continuum robots, dexterity primarily refers to the accessibility of the robot's end-orientation and the number of redundant solutions for each operational position—a challenge known as the inverse kinematics problem. Deriving an analytical solution for continuum robots, which exhibit multi-joint coupling and an infinite number of degrees of freedom, is highly complex \cite{Zhang23}. Consequently, numerous researchers have concentrated on studying the inverse kinematics of these robots. While the Newton-Raphson iteration method \cite{Singh17} can enhance accuracy, it suffers from low computational efficiency. Similarly, the Jacobian matrix approach for solving differential inverse kinematics \cite{xu2010} is both inefficient and susceptible to singularities. To address these issues, the real-time numerical integration technique \cite{Bieze18} employs Lagrange multiplier-based numerical optimization within the finite element method to solve the inverse kinematics problem.

Considering the system's subjection to the constraint of the remote center of motion (RCM) is a key factor in every robot-assisted MIS \cite{Azimian2010,Aghakhani2013,sadeqian2019,Yoshida2018,Zhang2021_RCM_cntum,zhao2025development,Wang2023Laparoscope}. Enhancing distal dexterity control and analysis of continuum instrument in minimally invasive surgeries (MIS) is crucial for performing complex tasks in confined environments and has been a major research focus in recent years \cite{Wang21,Ai24,Zhang22}.  In researches such as \cite{Zhang2020,Wang2023_endoscop} authors introduces an innovative flexible robotic endoscope, featuring a constrained tendon-driven continuum mechanism , specifically designed for applications in bariatric surgery.
Other research \cite{Razjigaev2021} focused on teleoperating a custom-designed snake-like manipulator for use in fiber-optic knee arthroscopy, optimizing it for dexterous manipulation.
In \cite{Cheng2024}, the authors introduce a spatial solid angle concept to assess the dexterity of continuum robots used in head and neck oncology, while Wang et al. \cite{Wang21} analyze the dexterity and performance of multi-segment continuum robots based on the robot's Jacobian. In \cite{Habich2024}, the authors introduce a shape-fitting approach to solve the inverse kinematics for a hyper-redundant continuum robot performance in a telemanipulation strategy.\par
%
 Research such as \cite{Rayne2018} explores control strategies for redundant continuum instruments in the context of minimally invasive surgery (MIS). They investigate various inverse kinematics (IK) control methods and evaluate them through a simulation-based user study, providing insights into the most suitable approaches for surgical teleoperation. Similarly, studies like \cite{Kwok2013} explore dimensionality reduction, real-time GPU-based proximity queries, and haptic feedback to enable robots to navigate safely and efficiently in endoscopy while minimizing computational demands.
On the other hand, research such as \cite{PALMER2014,Lilj2014} focuses on the motion planning and navigation of continuum instruments. These methods eliminate the need for predefined environment models. Additionally, by shifting from joint-space control to shape-based planning, they enable continuum robots to autonomously generate complex 3D movement patterns with greater adaptability. 
However, none of these studies simultaneously consider all these aspects—RCM constraint, continuum instrument performance investigation, and teleoperation framework—together in a unified control approach. \par
Most researchers, as mentioned above, primarily focus on the performance and dexterity of continuum instruments without accounting for constraints such as the RCM—a fundamental aspect of all robot-assisted MIS—nor the intrinsic coupling of linear and angular commands in continuum instruments. Therefore, investigating teleoperated linear and angular commands of continuum instruments under constraints like the RCM is crucial and, to the best of our knowledge, has not been fully explored in the literature. 
In this work, we focus on the application space of laparoscopic continuum instruments with an emphasis on the kinematics to support teleoperation. The remaining challenges are multifaceted: (1) linear and angular motions are inherently coupled in the continuum instrument tips compared to their counterpart (e.g. dVRK wristed instruments); (2) common haptic devices used in teleoperation has easy decoupled linear and angular inputs and (3) the RCM kinematic constraint complicates the conflict between the aforementioned (1) and (2). \par
The contributions of this research paper are summarized as follows:
\begin{itemize}
\item [1.] {Quantitatively investigating the kinematic performance of continuum instrument during teleoperation under RCM constraints, comparing linear and angular commands using a task-priority approach.}
\item[2.] {Developed and adopted a novel task-priority-based kinematic formulation for the quantitative analysis and comparison of teleoperation commands for continuum instruments.}
\item [3.] {Experimental development that includes designing an instrument module (IM) and integrating it with a 7-DoF manipulator. The teleoperation setup incorporates a ROS1-ROS2 bridge to enable communication between the ROS1-based haptic stylus and the main motion controller in ROS2, along with a Micro-ROS platform for real-time servo control.}
\end{itemize}\par

\section{KINEMATICS, DESIGN AND MODELING}
In this section, we discuss the kinematics to support the laparoscopic continuum instrument position and orientation control. As shown in figures\ref{fig:System_arch} and~\ref{fig:Robotic_system}, a laparoscopic continuum instrument module is mounted on a 7-DoF robot manipulator. The following sections discuss the kinematics of the continuum instrument tip, the kinematics of the robot manipulator, and the instantaneous kinematics of the integrated system. For clarity, we have adopted the nomenclature presented in Table \ref{table:nomencluture}.

\subsection{Robotic Continuum Instrument}
\begin{figure}[ht]
    \centering
        \includegraphics[width=1\linewidth]{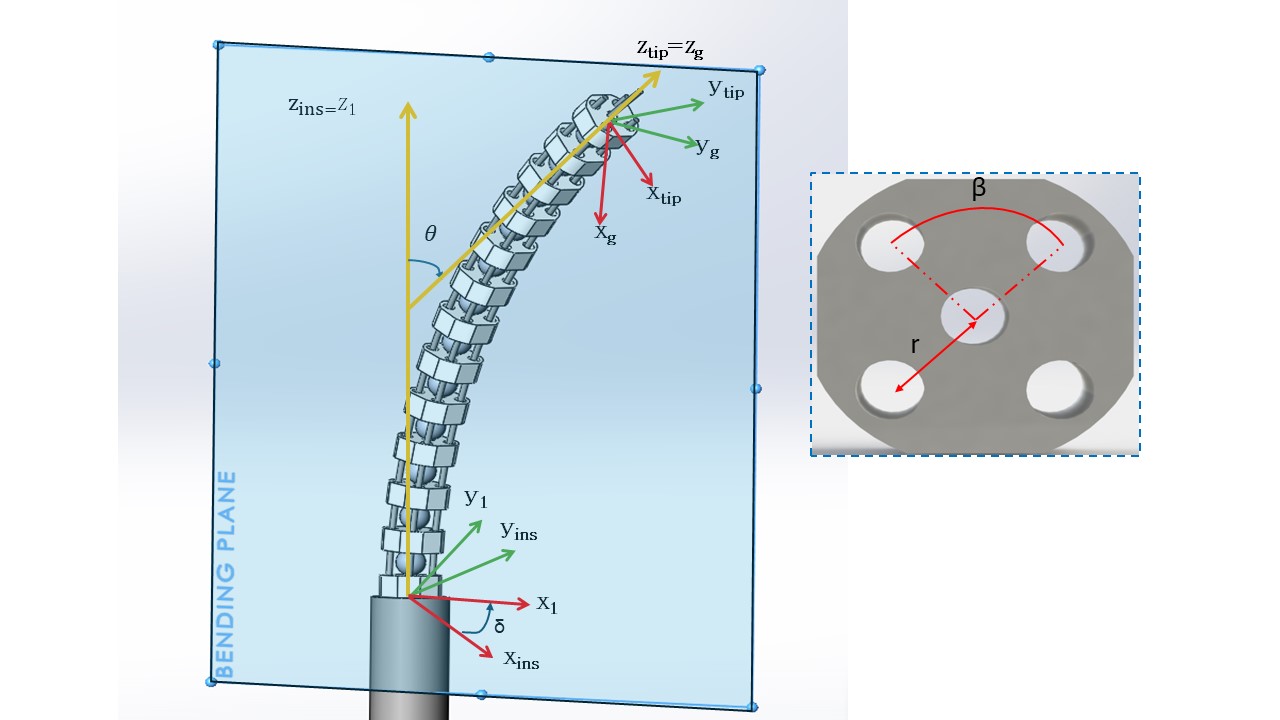}
    \caption{Kinematics of the continuum instrument and the bending plane}
    \label{conintuum_arm}
\end{figure}
In this section, we analyze the kinematics of a four-tendon-driven continuum instrument to determine the relationship between the position and orientation of the robot's tip and the pulling length of each wire. Figure.\ref{conintuum_arm} illustrates the schematic kinematics of the flexible instrument. Typically, the four wires are attached to the robot's tip, and their manipulation causes the continuum instrument to bend in specific directions. The kinematic configuration of the robot is precisely described using two angles: one representing the bending direction and the other the magnitude of bending. These angles are denoted as $\theta$ and $\delta$ , respectively. The pulling lengths of the four wires and the relative position/orientation of the instrument's tip can be concisely expressed using a homogeneous transformation matrix, denoted as ${^\text{b} \mathrm{T}}_{\text{tip}}$. These parameters are functions of $\theta$ and $\delta$. Determining these functions is essential to resolving the kinematic relationships. The calculation of the pulling lengths is outlined as follows:
\begin{align}
    \bs{\ell}&= \mathbf{f}_\mathbf{\ell}(\bs{\psi}), \quad \bs{\ell}
    = \begin{bmatrix} 
        l_1 & l_2 & l_3 & l_4 
    \end{bmatrix}^\mathrm{T}\\
    \bs{\psi} &= \begin{bmatrix}
        \theta & \delta
    \end{bmatrix}^\mathrm{T} 
    \label{eq:1}
\end{align}
The length of the ith tendon is given by,
\begin{equation}
L_i = L + \ell_i 
\end{equation}
The relationships between the configuration space parameters and the pulling lengths, denoted as \( \ell \), are as follows:
\begin{equation}
\delta = \delta_1 = \text{atan2}\left(l_2 - l_1 \cos \beta, -l_1 \sin \beta\right)
\end{equation}
where, 
\begin{equation}
r_i \triangleq r \cos(\theta_i) \quad \text{for } i = 1, 2, 3, \text{ and } 4
\end{equation}
\begin{equation}
\delta_i \triangleq  \delta + (i - 1) \theta \quad \text{for } i = 1, 2, 3, \text{ and } 4
\end{equation}
The instantaneous inverse kinematics is given as,
\begin{equation}
\dot{\bs{\ell}} = \mb{J}_{\mb{\ell}\psi} \dot{\bs{\psi}}, \qquad \mb{J}_{\mb{\ell}\psi} \in \mathbb{R}^{4\times2} \label{eq:actutation_jac}
\end{equation}
\begin{table} 
    \centering
    \caption{Nomenclature for Kinematics }
    \label{tab:nomenclature}
    {\renewcommand{\arraystretch}{1.5}
        \footnotesize
 \begin{tabular}{m{.15\columnwidth} m{.65\columnwidth}}
            \hline
            Symbol & \vspace{1mm} Description \vspace{1mm} \\
            \hline
            \({x_1,y_1,z_1}\) & Bending plane coordinate system \\
            \({{x_{\text{ins}}},y_{\text{ins}},z_{\text{ins}}}\) & Continuum instrument base disk  coordinate system \\
            \({x_{\text{tip}},y_{\text{tip}},z_{\text{tip}}}\) & Continuum instrument end disk coordinate system \\
             \({x_g,y_g,z_g}\) & An imaginary gripper coordinate system attached to the end disk \\
            \(i\) & Index of the secondary backbones, \(i = 1,2,3,4\) \\
            \(L, L_i\) & Length of the primary and the \(i\)th secondary backbone measured from the base disk to the end disk \\
            \(\ell_i\) & Joint parameter of the \(i\)th secondary backbone \(\ell_i = L_i - L\) \\
            \(r\) & Radius of the pitch circle defining the positions of the secondary backbones in all the disks. \\
            \(\beta\) & Division angle of the secondary backbones along the circumference of the pitch circle, \(\beta = \frac{2\pi}{4}\) \\
            \(\theta_i\) & The angle of the tangent to the primary backbone in the bending plane.  Note that  \( \theta_0 = \frac{\pi}{2} \) is a constant.   \\
            \(\delta_i\) & A right-handed rotation angle from \( {x}_1 \) about \( {z}_1 \) to a line passing through the primary backbone and the \(i\)th secondary backbone. At a straight configuration \( {x}_1 \) is along the same direction as the desired instantaneous linear velocity of the end disk. \\
            \(\delta\) & \( \delta = \delta_1 \) and \( \delta_i = \delta + (i - 1)\beta \), \;\(i = 1,2,3,4\) \\
            \(\Delta_i\) & Radial offset from primary backbone to the projection of the \(i\)th secondary backbone on the bending plane. \\
            \(\mathbf{J}_{yx}\) & Jacobian matrix of the mapping \( \dot{y} = \mathbf{J}_{yx} \dot{x} \) where the dot over the variable represents time derivative. \\
            \(^1\mathbf{R}_2\) & Rotation matrix of frame 2 with respect to frame 1. \\
            \( ^\mb{1}\mb{p}\) & Position vector of the continuum instrument tip in bending plane frame \\
            \( \lambda\) & The interpolation variable of RCM point along the surgical instrument shaft, $\lambda \in(0,1)$ \\
            \({\mathbf{x}}_{\text{tip}}\) & The continuum instrument's tip pose,  \({\mb{x}} \in \mathbb{R}^{4 \times 4} \)\\
            \hline
 \end{tabular}
    }
    \label{table:nomencluture}
\end{table}
\begin{equation}
\mathbf{J}_{\ell\psi} \triangleq 
\left[\begin{array}{c:c}
\mathbf{J}_{\ell\psi_\theta} & \mathbf{J}_{\ell\psi_\delta}
\end{array}\right] = 
\left[\begin{array}{c:c}
r\mb{c}\delta_1 & -r\theta_l \mb{s}{\delta_1} \\
r\mb{c}\delta_2 & -r\theta_l  \mb{s}{\delta_2} \\
r\mb{c}\delta_3 & -r\theta_l  \mb{s}{\delta_3} \\
r\mb{c}\delta_4 & -r\theta_l  \mb{s}{\delta_4}
\end{array}\right] \label{eq:Actuatation_to_Task_Mapping}
\end{equation}
where,
\begin{align}
    \mb{s}{\delta_i} = \text{sin} \hspace{2pt}{\delta_i} \\
    \mb{c}{\delta_i} = \text{cos} \hspace{2pt}{\delta_i}
\end{align}
The \(\mb{J}_{\mb{x}{\psi}}\) is formulates as follow:
\begin{equation}
    \mathbf{J}_{\mb{x}{\psi}} = 
    \left[ 
    \begin{array}{c}
    \mathbf{J}_{\mb{p}\hspace{1pt}\psi} \\ 
    \mathbf{J}_{\bs{\omega}\hspace{1pt}\psi} 
    \end{array} 
    \right] \label{eq:ins_jac}
\end{equation}
The instantaneous direct kinematics of the tip with respect to the base of the continuum segment can be expressed as,
\begin{equation}
{\bs{\xi}}_{\text{tip/ins}} = \mb{J}_{\mb{x}{\psi}} \dot{\bs{\psi}} ,\qquad \mb{J}_{\mb{x}{\psi}} \in \mathbb{R}^{6\times2}
\label{eq:map_joint_to_task}
\end{equation}
The Jacobian matrix \( \mathbf{J}_{\mb{p}\hspace{1pt}\psi} \) in \eqref{eq:ins_jac} can be obtained by taking the time derivative of \( {^{\text{ins}}\mathbf{\mb{p}}} \):
\begin{equation}
   { ^{\text{ins}}\mathbf{\mb{p}}} = ^{\text{ins}}{\mb{R}_1} ^1{\mb{p}}
\end{equation}
\begin{equation}
^1\mathbf{\mb{p}} = \frac{l}{-(\theta_0-\theta_l)} 
\begin{bmatrix}
s \theta_l -s \theta_0 & \;0 &\; c\theta_0 - c\theta_l
\end{bmatrix}^\mathrm{T}
\end{equation}
where, referring to Fig.\ref{conintuum_arm}, \(^1\mathbf{\mb{p}}\)  is the position of the tip of the continuum instrument in
the \textit{bending plane} coordinate,
\begin{equation}
\mathbf{J}_{\mb{p \psi}} = L
\begin{bmatrix}
\frac{(\theta_l - \theta_0) c_{\theta_l} - s_{\theta_l} + 1}{(\theta_l - \theta_0)^2} & c_{\theta_l} \frac{s_{\theta_l}(\theta_L - 1)}{\theta_l - \theta_0} \\
- s_{\theta_l} \frac{(\theta_l - \theta_0) c_{\theta_l} - s_{\theta_l} + 1}{(\theta_l - \theta_0)^2} & -c_{\theta_l} \frac{c_{\theta_l}(s_{\theta_l} - 1)}{\theta_l - \theta_0} \\
\frac{(\theta_l - \theta_0) s_{\theta_l} + c_{\theta_l}}{(\theta_l - \theta_0)^2} & 0
\end{bmatrix}
\end{equation}

\begin{equation}
^{\text{ins}}\bs{\omega}_g = -\dot{\delta} {\hat{\mb{z}}}_{\text{ins}} + ^{\text{ins}}\mb{R}_1 ((-\dot{\theta}_L ^1 \hat{\mb{y}}_1) + ^1\mb{R}_{\text{tip}}(\dot{\mb{\delta}}^{\text{tip}} \hat{\mb{z}}_{\text{tip}}))
\end{equation}
\begin{equation}
\mathbf{J}_{\bs{\omega}{\psi}} = 
\begin{bmatrix}
-s_{\delta} & c_{\delta}c_{\theta_l} \\
-c_{\delta} & -s_{\delta}c_{\theta_l} \\
0 & -1 + s_{\theta_l}
\end{bmatrix}
\end{equation}
\subsection{Kinematics of the Manipulator}
\begin{figure}[ht]
    \centering
    \includegraphics[width=\columnwidth] {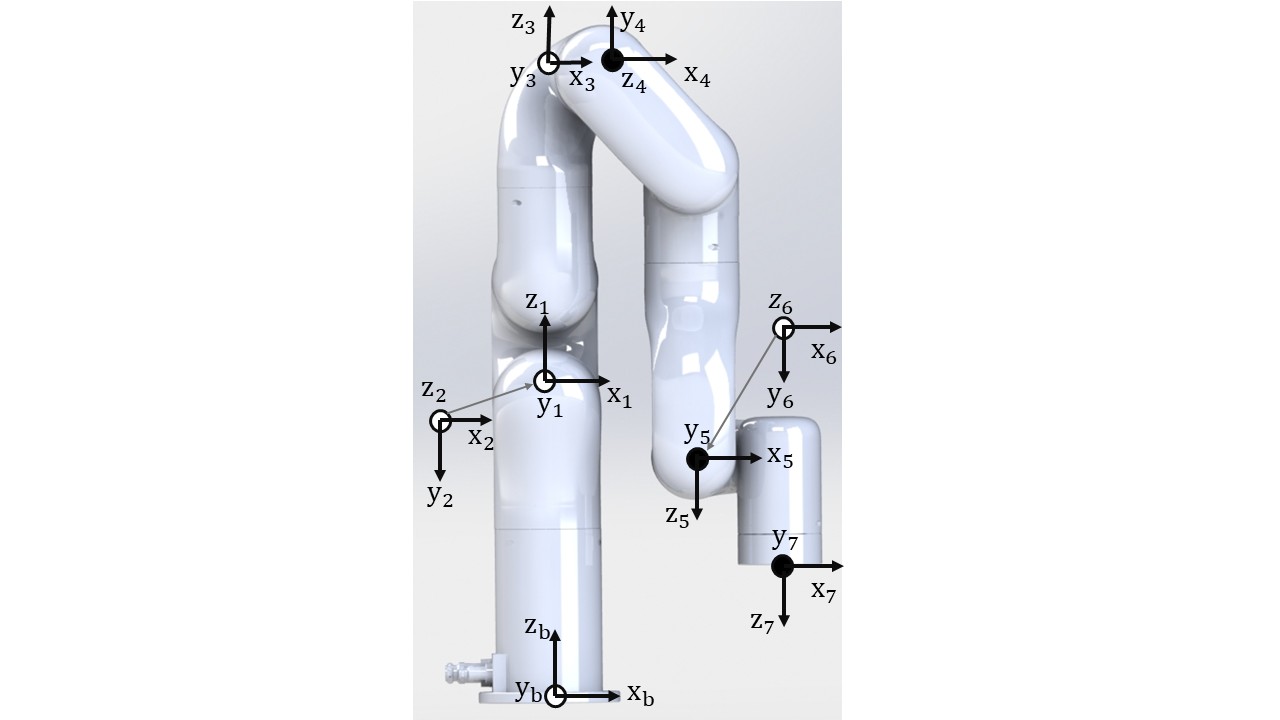}
    \caption{ The manipulator reference coordinates } 
    \label{fig:Robotic_system_coord}
\end{figure}

Figure.\ref{fig:Robotic_system_coord}  shows the manipulator in its home configuration. We examine the direct kinematics from joint space to task space. The task space is characterized by the position and orientation of frame \{7\}, as described by the homogeneous transformation \(^{\text{b}}\mb{T}_{7} \).
\begin{equation}
^\text{b}\mb{T}_{7} = 
\begin{bmatrix} \label{transformation_matrix}
^\text{b}\mb{R}_{7} & ^\text{b}\mb{p}_{7} \\
0 & 1
\end{bmatrix}
\end{equation}
where \(^\text{b}\mb{p}_{7}\) is the position of the origin of frame {7} and \(^\text{b}\mb{R}_{7}\) is the orientation of frame, relative to the robot base \emph{\{b\}}.
we define the configuration space \({\mb{q}}_{\text{arm}}\), to be \(\mb{q}_{\text{arm}} = [ q_1,q_2,\cdot \cdot \cdot, q_7]^\mathrm{T}\), when we don't have the continuum instrument attached to the wrist.
\begin{table}[h]
\caption{The Manipulator Kinematic Parameters at Home Configuration}
\centering
\begin{tabular}{|p{1.6cm}|p{0.8cm}|p{0.8cm}|p{0.8cm}|p{0.8cm}|p{0.8cm}|}
\hline
\textbf{Kinematics} & \textbf{theta (rad)} & \textbf{d (mm)} & \textbf{alpha (rad)} & \textbf{a (mm)} & \textbf{offset (rad)} \\
\hline
Joint1 & 0 & 267 & 0 & 0 & 0 \\
\hline
Joint2 & 0 & 0 & -$\pi/2$ & 0 & 0 \\
\hline
Joint3 & 0 & 293 & $\pi/2$ & 0 & 0 \\
\hline
Joint4 & 0 & 0 & $\pi/2$ & 52.5 & 0 \\
\hline
Joint5 & 0 & 342.5 & $\pi/2$ & 77.5 & 0 \\
\hline
Joint6 & 0 & 0 & $\pi/2$ & 0 & 0 \\
\hline
Joint7 & 0 & 97 & -$\pi/2$ & 76 & 0 \\
\hline
\end{tabular}
\label{table:kinematics}
\end{table}
The direct kinematics can be represented through a sequence of moving frames by multiplying the link parameter matrices in the following manner:
\begin{equation}
    ^\text{b}\mb{T}_{7} = \prod_{k=1}^{7} \; ^{\text{k-1}}\mb{T}_{\text{k}}
    \label{direct_kin}
\end{equation}
where, \(^{\text{k-1}}\mb{T}_{\text{k}}\) is the transformation matrix relating the \(\text{k}^\text{th}\) coordinate frame to the \(\text{(k-i)}^\text{th}\) coordinate frame, for the given sets of properties mentioned in Table~\ref{table:kinematics}:
\begin{equation}
^{\text{k-1}}\mb{T}_{\text{k}} = \begin{bmatrix}
c\theta_i & -c\alpha_i s\theta_i & s\alpha_i s\theta_i & a_i c\theta_i \\
s\theta_i & c\theta_i c\alpha_i & -s\alpha_i c\theta_i & a_i s\theta_i \\
0 & s\alpha_i & c\alpha_i & d_i \\
0 & 0 & 0 & 1 
\end{bmatrix}
\end{equation}
\par
Considering \eqref{direct_kin}, we map the joint velocities to the end-effector twist \({\bs{\xi}}_{\text{ee}}\) using Jacobian, \(\mb{J}_{\text{ee}}\),
\begin{equation}
    {\bs{\xi}}_{\text{ee}} = \mb{J}_{\text{ee}}(\mb{q}_{\text{arm}}) \dot{\mb{q}}_{\text{arm}}, \qquad   \mb{J}_{\text{ee}} \in \mb{R}^{6 \times 7}
    \label{end_kin}
\end{equation}
 the end-effector Jacobian \(\mb{J}_{\text{ee}}\), is given by the Plücker line coordinates of each joint axis as expressed in the base frame and centered at the origin
of frame {7} or task-space frame:
\begin{equation}
(\mb{J}_{\text{ee}}(\mb{q}_{\text{arm}}))^\text{k} = 
\begin{cases} 
\begin{pmatrix}
\hat{\mathbf{z}}_{\text{k-1}} \times (\mathbf{o}_7 - \mathbf{o}_{\text{k-1}}) \\
\hat{\mathbf{z}}_{\text{k-1}}
\end{pmatrix}
, & \text{if revolute joint} \\[12pt]
\begin{pmatrix}
\hat{\mathbf{z}}_{\text{k-1}} \\
0
\end{pmatrix}
, & \text{if prismatic joint}
\end{cases}
\end{equation}
where \( \mb{o}_{k} \), denotes the position of frame \{k\}. In addition, \emph{\(\text{\{ee\}}\)} here is equivalent to frame {7} in Fig.\ref{fig:Robotic_system_coord}. \par
\subsection{Remote Center of Motion (RCM) Constraint Kinematics}
\begin{figure}[t!]
    \centering
    \includegraphics[width=0.9\columnwidth, height=0.5\textheight, trim={4.0cm 0cm 4.5cm 0cm}, clip]{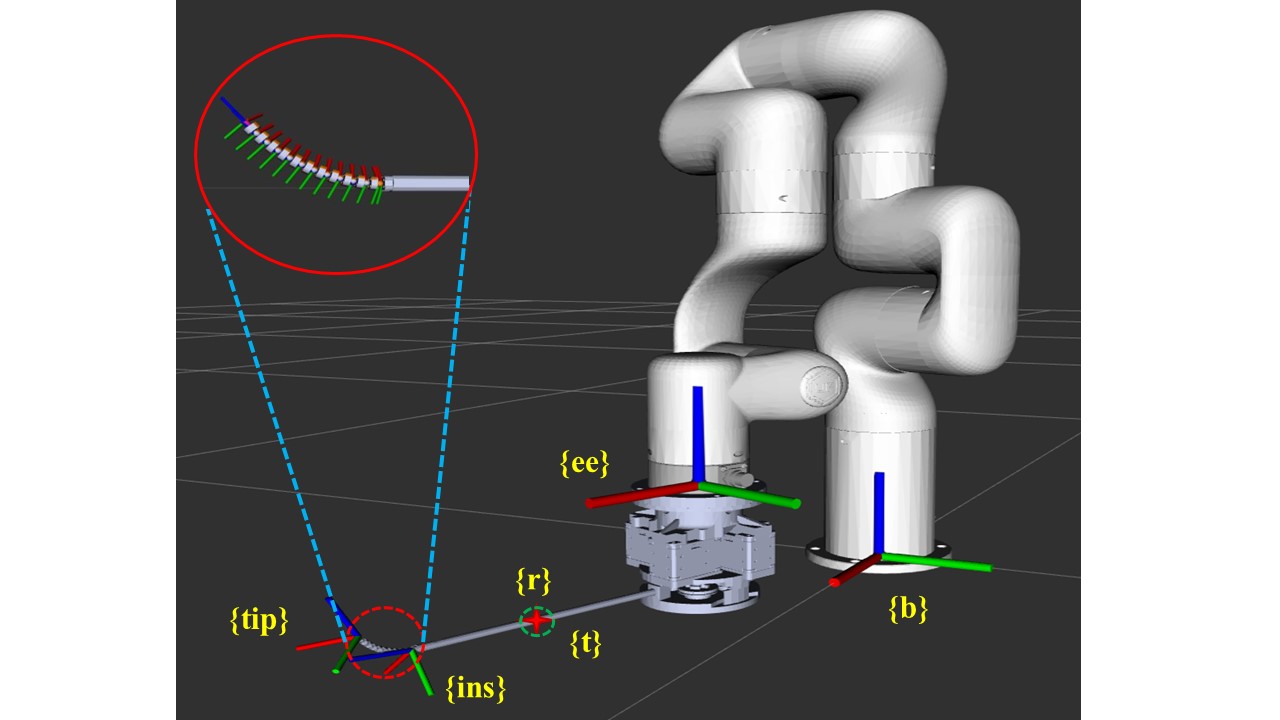}
    \caption{The robotic system, includes 7-DoF manipulator and a custom-designed instrument module}
    \label{fig:Robotic_system}
\end{figure}

Referring to the system depicted in Fig.\ref{fig:Robotic_system}, the incision point \( {r} \) imposes remote center of motion (RCM) kinematic constraint on the surgical instrument shaft of the IM. Let \(r \) represent the point that is momentarily coincident with trocar (point \( {t} \)). The RCM constraint at the trocar permit sliding along and rolling about the shaft's longitudinal axis, along with two tilting motions, while ensuring the shaft's longitudinal axis remains coincident with trocar.
This RCM constraint require that the velocity of point \( r \) in a plane perpendicular to the shaft's axis must match the velocity of the trocar point \({t} \). 
It is evident that if we assume trocar is fixed, the RCM velocity becomes zero. On the other hand, if we consider the trocar velocity obtained from a sensor or estimate it using the method presented in \cite{nasiri2024Admittance_RCM,nasiri2024teleoperation}, the robot should do the task while respecting he RCM constraint.

\begin{figure}[ht]
    \centering
    \includegraphics[width=\columnwidth, height=0.5\columnwidth, trim=0 0 0 6, clip]{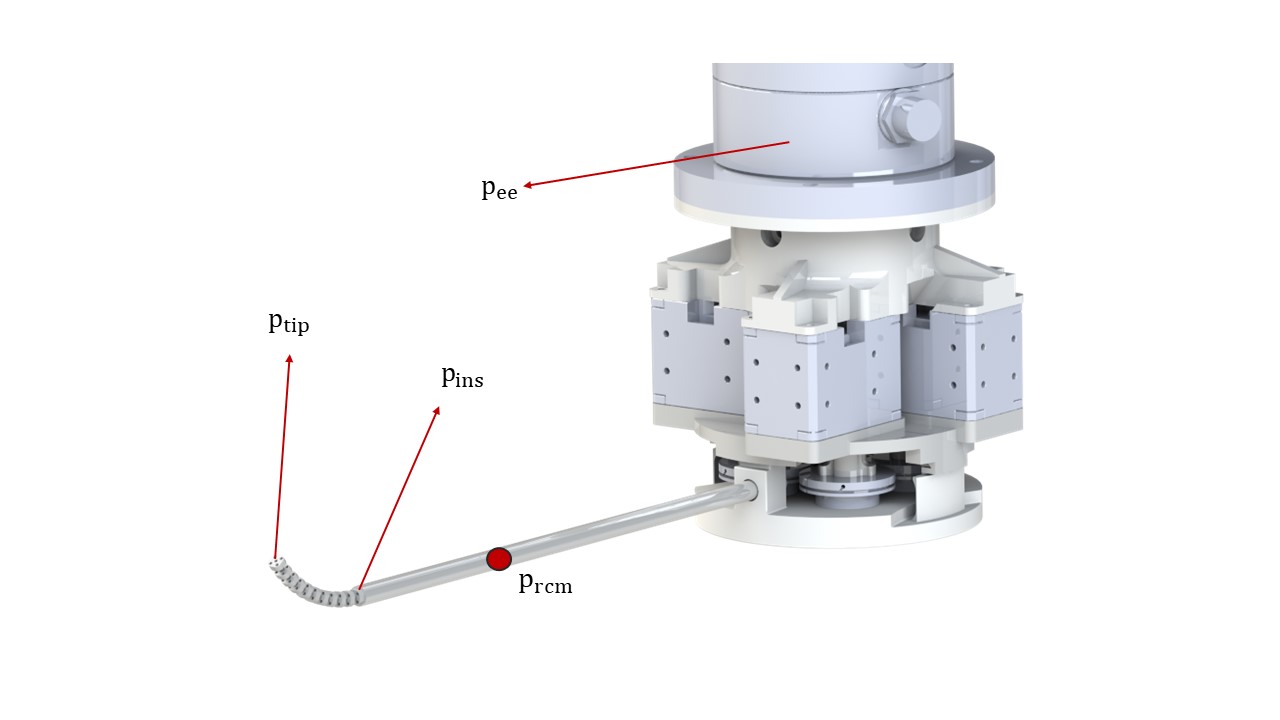}
    \vspace{-30pt}
    \caption{Schematic of a custom-designed, miniature tendon-driven continuum surgical instrument}
    \label{fig:custom_ins_rcm}
    \vspace{-15pt}
\end{figure}

By defining the linear velocity  of \( \text{RCM} \) as \( {\bs{\xi}}_{\text{rcm}} \), the RCM kinematic constraint can be expressed as follows, referring to Fig.\ref{fig:custom_ins_rcm}:
\begin{align}
 &\mb{p}_{\text{rcm}} = \mb{p}_{\text{ee}} + \lambda (\mb{p}_\text{ins} - \mb{p}_{\text{ee}}), \quad \lambda \in (0,1), \; \mb{p}_{\text{rcm}}\in \mathbb{R}^{6 \times 1} \label{eq:def_RCM} \\
 & \mb{d}_\text{ins} = \mb{p}_\text{ins} - \mb{p}_\text{\text{ee}}  \\
& \mb{v}_{\text{rcm}} = 
\begin{bmatrix}
\begin{array}{c;{2pt/2pt}c}
(\mb{J}_{\text{ee}} + \lambda (\mb{J}_\text{ins} - \mb{J}_{\text{ee}})) & \mb{d}_\text{ins}
\end{array}
 \end{bmatrix}
\underbrace{\begin{bmatrix}
\dot{\mb{q}}_{\text{arm}} \\ \dot{\lambda}
\end{bmatrix}}_{ \triangleq \; \dot{\mb{q}}_{\text{rcm}}}
\label{eq:rcm_veloc}
\end{align} 
The RCM constraint linear velocity \({\bs{\xi}}_{\text{rcm}}\), is then derived from \eqref{eq:rcm_veloc},
\begin{align}
& {\bs{\xi}}_{\text{rcm}} = \mb{s}_m \; \mb{v}_{\text{rcm}}, \qquad {\bs{\xi}}_{\text{rcm}} \in \mathbb{R}^{3 \times 1} \label{rcm_vel} \\
& \mb{s}_m = 
    \begin{bmatrix}
    \begin{array}{c;{2pt/2pt}c}
      \mb{I} &  \mb{0} 
    \end{array} 
    \end{bmatrix}, \qquad \mb{s}_m \in   \mathbb{R}^{3 \times 6}
\end{align}
where, \(\mb{s}_m\) is the selection matrix to extract the linear velocity of the \emph{RCM}, from \(\mb{v}_{\text{rcm}}\) twist. 
Substituting \eqref{eq:rcm_veloc} into \eqref{rcm_vel},
\begin{align}
   & {\bs{\xi}}_{\text{rcm}} =  \mb{J}_{{\text{rcm}}}(\mb{q}_{{\text{arm}}}, \lambda) \;
        \mb{\dot{q}}_{{\text{rcm}}}
    , \qquad   {\bs{\xi}}_{{\text{rcm}}} \in \mb{R}^{3 \times 1} \label{eq:rcm_velocity} \\
   & \mb{J}_{{\text{rcm}}} =\mb{s}_m \; \begin{bmatrix}
    \begin{array}{c;{2pt/2pt}c}
        (\mb{J}_{\text{ee}} + \lambda (\mb{J}_\text{ins} - \mb{J}_{\text{ee}})) & \mb{d}_\text{ins}
    \end{array}
    \end{bmatrix},\quad   \mb{J}_{{\text{rcm}}} \in \mb{R}^{3 \times 8}
    \label{eq:rcm_final_velocity}
\end{align}
Referring to \eqref{end_kin}, we can have the \emph{ins} twist, as follow:\\
\begin{equation}
    \bs{\xi}_{\text{ins}} = \mb{J}_{{\text{ins}}}(\mb{q}_{{\text{arm}}}) \dot{{\mb{q}}}_{{\text{arm}}}, \quad   \mb{J}_{{\text{ins}}} \in \mathbb{R}^{6 \times 7}
    \label{eq:inst_kin}
\end{equation}
Considering Fig.\ref{fig:Robotic_system} and Fig.\ref{fig:custom_ins_rcm}, we can now derive the velocity of the continuum instrument \emph{tip}, accordingly.
\begin{equation}
    {\bs{\xi}}_{\text{tip}} =  \bs{\xi}_{\text{ins}} + {\bs{\xi}}_{\text{tip\hspace{1pt}/ins}}
    \label{eq:tip_velocity}
\end{equation}

The first term in \eqref{eq:tip_velocity} is the continuum segment base(ins) twist and the second term is define the relative twist of \emph{tip} with respect to the \emph{ins}. 
The instrument base twist \({\bs{\xi}}_{\text{ins}}\) in \eqref{eq:tip_velocity} is defined in \eqref{eq:inst_kin}. The relative twist of the tip, previously mentioned in \eqref{eq:map_joint_to_task},  can be redefined as being determined by the motion of frame {ins} with the continuum segment held stationary, combined with the motion contribution from the continuum segment itself:
\begin{align}
&{\bs{\xi}}_{\text{tip\hspace{1pt}/ins}} = \mb{s}_{\text{ins}} \; {\bs{\xi}}_{\text{ins}}
  + 
 \mb{A}_\text{adj} \; \mb{J}_{\mb{x\psi}} \dot{\bs{\psi}}
 \label{eq:relative_twist} \\
 & \mb{s}_{\text{ins}}= \begin{bmatrix}
 \mb{0} & [{{\mb{n}}_{\text{tip\hspace{1pt}/ins}}}]_{\times}  \\
 \mb{0} & \mb{0}
\end{bmatrix}, \quad \mb{s}_{\text{ins}} \in \mathbb{R}^{6\times 6} \\
 & \mb{A}_\text{adj} = \begin{bmatrix}
\mb{R}_\text{ins} & \mb{0} \\
\mb{0} & \mb{R}_\text{ins} 
\end{bmatrix},  \quad  \mb{A}_\text{adj} \in \mathbb{R}^{6\times 6} \\
& {{\mb{n}}_{\text{tip\hspace{1pt}/ins}}} = \mb{s}_m (\mb{p}_{\text{ins}} - \mb{p}_{\text{tip}} )
\end{align}
Where, $[.]_\times$ represents the Skew-symmetric matrix operator and \(\mb{s}_{\text{ins}}\) is a selection matrix that extracts the angular velocity from the ins twist. The first term in \eqref{eq:relative_twist} is the velocity contribution of \emph{ins} frame, due to rotational velocity of the manipulator \(\bs{\omega}_{\text{ins}}\), acting on hypothetical link 
 connected ins to tip. The second term in \eqref{eq:relative_twist}, as referenced in \eqref{eq:map_joint_to_task}, represents the relative twist of the continuum instrument tip, caused by the motion of the continuum segment while the manipulator remains stationary. The \(\mb{A}_\text{adj}\) is an adjoint matrix that express the continuum segment \emph{tip} twist in base frame.\\ 
Then substituting \eqref{eq:inst_kin} in \eqref{eq:relative_twist},
\begin{align}
{\bs{\xi}}_{\text{tip\hspace{1pt}/ins}} = 
 \begin{bmatrix}
 \mb{0} & [{{\mb{n}}_{\text{tip\hspace{1pt}/ins}}}]_{\times}  \\
 \mb{0} & \mb{0}
\end{bmatrix}  \mb{J}_{\text{ins}}(\mb{q}_{\text{arm}}) \dot{{\mb{q}}}_{\text{arm}} + \mb{A}_\text{adj} \; \mb{J}_{\mb{x\psi}} \dot{\bs{\psi}}
\end{align} 
Now, we can rewrite the continuum instrument \emph{tip} twist \eqref{eq:tip_velocity}, as follow:
\begin{align}
& {\bs{\xi}}_{\text{tip}} =  \mb{J}_{\text{ins}}(\mb{q}_{\text{arm}}) \dot{\mb{q}}_{\text{arm}} 
+ \begin{bmatrix}
 \mb{0} & [{{\mb{n}}_{\text{tip\hspace{1pt}/ins}}}]_{\times}  \\
 \mb{0} & \mb{0}
\end{bmatrix}  \mb{J}_{\text{ins}}(\mb{q}_{\text{arm}})\; \dot{{\mb{q}}}_{\text{arm}} + \mb{A}_\text{adj} \mb{J}_{\mb{x\psi}} \dot{\bs{\psi}}\\
& {\bs{\xi}}_{\text{tip}} = \underbrace{ \left( \begin{bmatrix}
 \mb{I} & [{{\mb{n}}_{\text{tip\hspace{1pt}/ins}}}]_{\times}  \\
 \mb{0} & \mb{I}
\end{bmatrix} \mb{J}_{\text{ins}}(\mb{q}_{\text{arm}})  \right)}_{\triangleq \; \mb{J}_{\text{aux}}} \dot{\mb{q}}_{\text{arm}} + \mb{A}_\text{adj}\; \mb{J}_{\mb{x\psi}} \dot{\bs{\psi}} \label{eq:ins_tip_final_velocity} \\
& \mb{J}_{\text{aux}} = \begin{bmatrix}
 \mb{I} & [{{\mb{n}}_{\text{tip\hspace{1pt}/ins}}}]_{\times}  \\
 \mb{0} & \mb{I}
\end{bmatrix} \mb{J}_{\text{ins}}(\mb{q}_{\text{arm}})
\end{align}
Considering \eqref{eq:ins_tip_final_velocity}, we can define the Jacobian of tip to be as follows:
\begin{align}
    & {\bs{\xi}}_{\text{tip}} \triangleq \left[ 
\begin{array}{c:c}
\multicolumn{2}{c}{\mb{v}} \\ 
\multicolumn{2}{c}{\bs{\omega}}  
\end{array} 
\right]= \mb{J}_{\mathrm{T}}\; \dot{\mb{q}}_{\mathrm{T}} \label{eq:p_dot_tip_ins_base}\\
    & \mb{J}_{\mathrm{T}} = \begin{bmatrix}
    \begin{array}{c;{2pt/2pt}c}
        \mb{J}_{\text{aux}} & \mb{A}_\text{adj} \hspace{1pt}\mb{J}_{\mb{x\psi}}
    \end{array}
    \end{bmatrix}, \quad  \mb{J}_{\mathrm{T}} \in \mathbb{R}^{6 \times 9} \label{eq:ins_tip_final_jacob}\\
 & {\dot{\mb{q}}}_{\mathrm{T}} =  \left[ \dot{\mb{q}}^\mathrm{T}_{\text{arm}} ,  {\dot{\bs{\psi}}}^\mathrm{T} \right]^\mathrm{T} , \quad  \dot{\mb{q}}_{\text{T}} \in \mathbb{R}^{9 \times 1} \label{eq:aug_state}
\end{align}
With \eqref{eq:rcm_final_velocity} and \eqref{eq:ins_tip_final_velocity}, we define an augmented twist, \(\dot{\bs{\Phi}}_{\text{aug}}\) to be as follow:
\begin{equation}
    \dot{\bs{\Phi}}_{\text{aug}} =  \left[ 
\begin{array}{c:c}
\multicolumn{2}{c}{{\bs{\xi}}_{\text{tip}}} \\ 
\hdashline
\multicolumn{2}{c}{{\bs{\xi}}_{\text{rcm}}}  
\end{array} 
\right], \qquad  \dot{\bs{\Phi}}_{\text{aug}} \in \mathbb{R}^ {9 \times 1 } \label{eq:aug_twist}
\end{equation}
We can incorporate the linear constraints using an augmented Jacobian method as follows:
\begin{align}
    & \dot{\bs{\Phi}}_{\text{aug}} = \mb{J}_{{\text{aug}}} \; \dot{\mb{q}}_{{\text{aug}}}, \qquad \dot{\bs{\Phi}}_{\text{aug}} \in \mathbb{R}^{9\times 1} \label{eq:total_aug_equation}\\ 
    & \dot{\mb{q}}_{{\text{aug}}} = \left[ \dot{\mb{q}}^\mathrm{T}_{\text{arm}} , {\dot{\bs{\psi}}}^\mathrm{T}, \dot{\lambda} \right]^\mathrm{T} , \quad  \dot{\mb{q}}_{\text{aug}} \in \mathbb{R}^{10 \times 1} \label{eq:aug_joint_velocity}\\
    & \mb{J}_{{\text{aug}}} = \begin{bmatrix}
    \mb{J}_\mathrm{T} & \mathbf{0}_{6 \times 1} \\
     \hdashline
    \multicolumn{2}{c}{\mb{J'}_{\text{rcm}}}
\end{bmatrix}, \quad \mb{J}_{\text{{aug}}}\in\mathbb{R}^{9\times10} \label{eq:aug_jac}
\end{align} 
The general solution to \eqref{eq:total_aug_equation} is:
\begin{equation}
        \dot{\mb{q}}_{{\text{aug}}} \\
        = \mathbf{J}^{\dagger}_{\text{aug}} \hspace{2pt} \dot{\bs{\Phi}}_{\text{aug}} + \left(\mathbf{I}-\mathbf{J}^{\dagger}_{\text{aug}} \; \;\mathbf{J}_{\text{aug}}\right) \boldsymbol{\eta}, \quad 
        \bs{\eta}\in \mathbb{R}^{10\times1} 
    \label{eq:redundancy_res} 
\end{equation} 
We add two columns of zero to the RCM Jacobian \(\mb{J}_{\text{rcm}}\), in \eqref{eq:rcm_final_velocity}, to maintain consistency in \eqref{eq:total_aug_equation} by using an updated Jacobian for RCM denoted as \(\mb{J'}_{\text{rcm}}\). 
In addition, for the sake of the simplification, in \eqref{eq:aug_jac}, we added the zero column in the upper part of the \(\mb{J}_{\text{aug}}\), extending  \(\mb{J}_\mathrm{T}\).
\begin{equation}
    \mb{J'}_\mathrm{T} \triangleq \left[ 
    \begin{array}{c:c}
        \multicolumn{2}{c}{\mb{J}_{L}} \\ 
        \multicolumn{2}{c}{\mb{J}_{A}} 
    \end{array} 
    \right] = \begin{bmatrix}
        \mb{J}_\mathrm{T} & \mathbf{0}_{6 \times 1} 
    \end{bmatrix}, \qquad \mb{J'}_\mathrm{T} \in \mathbb{R}^{6\times10}
\end{equation}
where, \(\mb{J}_{L}\) and \(\mb{J}_{A}\) are the linear and angular part of the Jacobian, \(\mb{J'}_\mathrm{T}\).
\subsection{Task Priority Formulation}  \label{task_priroity}
We aim to investigate the kinematic performance of teleoperation commands under different task priorities. To achieve this, we use the task priority method in redundancy resolution, which ensures that multiple objectives in a redundant robotic system are managed hierarchically. This approach is supported by the literature, such as \cite{Baerlocher1998,Nakamura1987}. For this purpose, \eqref{eq:redundancy_res} can be considered as \textit{case0}, as there is no priority sequence for the system to follow. The system uses the augmented Jacobian approach to follow the teleoperated trajectory while respecting the RCM constraint.
In \textit{case0}, the linear and angular commands of the \textit{tip} are of equal importance. On the other hand, in the following cases, we developed and formulated a novel task-priority-based kinematics approach to investigate the teleoperation commands of a continuum instrument under RCM constraints. For instance, in \textit{case1}, we prioritize the linear command of the tip over its angular velocity.
\begin{itemize}
\item{Case 1:} \label{case1}
\begin{align}
&\text{Level 1, \textit{RCM}:} \quad \quad \mb{J'}_{{\text{rcm}}} \dot{\mb{q}}_{\text{aug}} = {\bs{\xi}}_{\text{rcm}}  \\%
& \text{Level 2, \textit{Linear}:} \quad \quad \mb{J}_{L} \dot{\mb{q}}_{{\text{aug}}} - {\mb{v}} \\
&\text{Level 3, \textit{Angular}: }  \quad \quad \mb{J}_{A} \dot{\mb{q}}_{{\text{aug}}} - {\bs{\omega}}  %
\label{eq:case1}
\end{align} 
\end{itemize}
The approach developed specifies the endeffector motion and satisfying the RCM constraint as a primary task while satisfying a minimizing the linear velocity of the continuum instrument tip. The mathematical proof of this task priority problem is explained in detail in Section.\ref{appdx1}.
The solution for \textit{case1} begins as follows:
\begin{align}
&\dot{\mb{q}}_{\text{aug},1} = \mb{J'}^\dagger_{\text{rcm}} {\bs{\xi}}_{\text{rcm}} + \mb{B}^\dagger \left({\mb{v}} - \mb{J}_{L} \mb{J'}^\dagger_{\text{rcm}} {\bs{\xi}}_{\text{rcm}}\right)  + \left(\mb{I} - \mb{J'}^\dagger_{\text{rcm}}  \mb{J}_{L} - \mb{B}^\dagger \mb{B}\right) \bs{\eta}
    \label{eq:final_redund_task_priority_form} \\
&\mb{B} = \mb{J}_{L} \left(\mb{I} -\mb{J'}^\dagger_{\text{rcm}} \mb{J'}_{\text{rcm}}\right)
\end{align}
where, \(\bs{\eta}\) can be be determined to satisfy a tertiary task. Following the same steps as above, solving for the \(\mb{\bs{\eta}}\), we will have the following final formulation for case1:
\begin{align}
& \dot{\mb{q}}_{\text{aug},1} = \mathbf{J'}_{\text{rcm}}^\dagger \, {\bs{\xi}}_{\text{rcm}} 
\,+\, \mb{B}^\dagger \left( {\mb{v}} \,-\, \ \mb{J}_{L} \mb{J}^\dagger_{\text{rcm}} {\bs{\xi}}_{\text{rcm}}\right)  + \left(\mb{I} - \mb{J'}^\dagger_{\text{rcm}}  \mb{J}_{L} - \mb{B}^\dagger \mb{B}\right) 
\mb{C}^\dagger \left( {\bs{\omega}} - \mathbf{b}_1 - \mathbf{b}_2 \right) + \mb{N}_3 \bs{\eta} \label{eq:final_task_pririty_case1}\\
& \mb{C} = \mb{J}_{A} \left(\mb{I} - \mb{J'}_{\text{rcm}}^\dagger \mathbf{J'}_{\text{rcm}} - \mb{B}^\dagger \mb{B} \right) \\
& \mb{b}_1 = \mb{J}_{A} \mb{J'}_{\text{rcm}}^\dagger  {\bs{\xi}}_{\text{rcm}} \\
& \mb{b}_2 =  \mb{J}_{A} \mb{B}^\dagger \left( {\mb{v}} -  \mb{J}_{L} \mb{J'}_{\text{rcm}}^\dagger {\bs{\xi}}_{\text{rcm}} \right) \label{eq:b2}\\
& \mb{N}_3 = \mathbf{I} - \mathbf{J'}_{\text{rcm}}^\dagger \mathbf{J'}_{\text{rcm}}^\dagger 
- \mb{B}^\dagger \mb{B} 
- \mb{D}\, \mb{C}^\dagger \mb{C} \\
&\mb{D} = \left( \mathbf{I} -  \mathbf{J'}_{\text{rcm}}^\dagger  \mathbf{J'}_{\text{rcm}}-\mb{B}^\dagger \mb{B}  \right)
\end{align}
In \eqref{eq:final_task_pririty_case1}, the arbitrary choice of \(\bs{\eta}\) will first satisfy the primary, then the secondary and finally the tertiary task.
\begin{itemize}
\item{Case 2:}
\begin{align}
&\text{Level 1, \textit{RCM}:} \quad \quad \mb{J'}_{{\text{rcm}}} \dot{\mb{q}}_{\text{aug}} = {\bs{\xi}}_{\text{rcm}}  \\%
& \text{Level 2, \textit{Angular}:} \quad \quad \mb{J}_{A} \dot{\mb{q}}_{{\text{aug}}} - {\bs{\omega}} \\
&\text{Level 3, \textit{Linear}:}  \quad \quad \mb{J}_{L} \dot{\mb{q}}_{{\text{aug}}} - {\mb{v}}  %
\label{eq:case2}
\end{align} 
\end{itemize}
In \textit{case2}, we consider the angular velocity command of the continuum instrument tip to be the second priority task, while the linear command is tertiary. We can follow the same procedure to formulate this task-priority case. In this scenario, the joint velocity commands are formulated to satisfy the RCM constraint, with priority given first to the angular twist command of the continuum, followed by the linear command from the haptic device.
\begin{align}
&\dot{\mathbf{q}}_{\text{aug},2} = \mathbf{J}_{\text{rcm}}^\dagger \, \bs{\xi}_{\text{rcm}} 
+ \mathbf{C'}^\dagger \left( \bs{\omega} - \mathbf{J}_{A} \mathbf{J}_{\text{rcm}}^\dagger  \bs{\xi}_{\text{rcm}} \right) + \left( \mathbf{I} - \mathbf{J}_{\text{rcm}}^\dagger \mathbf{J}_{A} - \mathbf{C'}^\dagger \mathbf{C'} \right) 
\mathbf{B'}^\dagger \left( \mathbf{v} - \mathbf{b'}_1 - \mathbf{b'}_2 \right) + \mathbf{N'}_3 \bs{\eta} \label{eq:final_task_pririty_case2}\\
&\mathbf{C'} = \mathbf{J}_A \left( \mathbf{I} - \mathbf{J'}_{\text{rcm}}^{\dagger} \mathbf{J'}_{\text{rcm}} \right)\\
&\mathbf{B'} = \mathbf{J}_L \left( \mathbf{I} - \mathbf{J'}_{\text{rcm}}^{\dagger} \mathbf{J'}_{\text{rcm}} - \mathbf{C'}^{\dagger} \mathbf{C'} \right)\\
&\mathbf{b'}_1 = \mathbf{J}_L \mathbf{J'}_{\text{rcm}}^{\dagger}  \bs{\xi}_{\text{rcm}} \\
&\mathbf{b'}_2 = \mathbf{J}_L \mathbf{C'}^{\dagger} \left( \bs{\omega} - \mathbf{J}_A \mathbf{J'}_{\text{rcm}}^{\dagger}  \bs{\xi}_{\text{rcm}} \right)\\
&\mathbf{N'}_3 = \mathbf{I} - \mathbf{J'}_{\text{rcm}}^{\dagger} \mathbf{J'}_{\text{rcm}} - \mathbf{C'}^{\dagger} \mathbf{C'} - \mb{D'}\, \mathbf{B'}^{\dagger} \mb{B'}\\
&\mb{D'} = \left( \mathbf{I} - \mathbf{J'}_{\text{rcm}}^{\dagger} \mathbf{J'}_{\text{rcm}}-\mathbf{C'}^{\dagger} \mathbf{C'}\right)
\end{align} 
We can use the augmented joint velocity commands in \eqref{eq:final_task_pririty_case1}, or~\eqref{eq:final_task_pririty_case2}, to simultaneously teleoperate the prioritized twist command while respecting the RCM constraint. Here is the definition of \(\bs{\eta}\) in the above equations:
\begin{align}
\bs{\eta} = \alpha \; \nabla(\mb{q},\lambda), \quad \; \alpha > 0, \; \alpha \in \mathbb{R}, \quad  \bs{\eta}\in\mathbb{R}^{10\times1} 
\label{eq:redundancy_res_modified} 
\end{align} \\
It is worth noting that the system have 1-DoF redundancy that can be resolved through an appropriate choice of \(\bs{\eta}\). Given that the manipulator has 7-DoF, the continuum instrument command imposes a 2-DoF motion constraint, and the RCM constraint adds another 2-DoF.
\begin{align}
 & \mb{g} = \frac{1}{2} \left\| {\lambda} - {\lambda}_0 \right\|^2 \\
 & \bs{\eta} = \alpha \; \nabla \mb{g}_{(\mb{q}, {\lambda})}  \bigg|_{{\lambda}_0 } \\ 
 & \bs{\eta} = \alpha \begin{bmatrix} \mb{0}_{9\times1} \\ \lambda - {\lambda}_0 \end{bmatrix}\label{eq:objective}
\end{align}
Finally, the joint commands for the manipulator and continuum segment can then be found,
\begin{equation}
    \mb{q}_{\text{aug}}= \mb{\dot{\mb{q}}}_{\text{aug}} \hspace{2pt} \text{dt} +\mb{q}_{\text{aug}} 
\end{equation} \par
\subsection{Kinematics of Instrument Driving Actuation Module}
To facilitate real hardware experiments, kinematics of instrument driving actuation module is needed. In this part, actuation module motor pulley positions are calculated based on the articulation (bending angle and bending plane) states. \par
Referring to \eqref{eq:actutation_jac} and recalling the following definitions, the actuation module's mapping to the task space can be derived,
\begin{align}
 \bs{\ell}_i &= \Delta_i  (\theta - \theta_0) \quad \text{for } i=1,\dots,4 \\
 \theta_0 &= \frac{\pi}{2} \\
 \beta &= \frac{2\pi}{4} = \frac{\pi}{2}
\end{align}
To express \(\dot{\bs{\ell}}_i\) in terms of \(\dot{\theta}\) and \(\dot{\delta}\), we differentiate \(\bs{\ell}_i\) with respect to time:
\begin{equation}
    \dot{\bs{\ell}}_i = \frac{d}{dt} (\Delta_i (\theta - \theta_0)) = \Delta_i \dot{\theta} + (\theta - \theta_0) \dot{\Delta}_i \label{eq:delta_dot}
\end{equation}
Since \(\Delta_i = r \cos(\delta_i)\), we can find the \(\dot{\Delta}_i\),
\begin{equation}
    \dot{\Delta}_i = -r \sin(\delta_i) \dot{\delta}_i
\end{equation}
Given \(\delta_i = \delta + (i-1)  \beta\) then,
\begin{equation}
    \dot{\delta}_i = \dot{\delta}
\end{equation}
Thus, \(\dot{\Delta}_i\) becomes:
\begin{equation}
    \dot{\Delta}_i = -r \sin(\delta + (i-1)  \beta) \dot{\delta}
\end{equation}
Substituting \(\dot{\Delta}_i\) back into \eqref{eq:delta_dot}:
\begin{equation}
    \dot{\bs{\ell}}_i = \Delta_i \dot{\theta} - r (\theta - \theta_0) \sin(\delta + (i-1)  \beta) \dot{\delta} \label{eq:86}
\end{equation}
Referring to \eqref{eq:actutation_jac}, the \eqref{eq:86} can be expressed in a matrix form as follows:
\begin{equation}
\underbrace{
\begin{bmatrix}
\dot{l}_1 \\
 \vdots \\
 \dot{l}_4 
\end{bmatrix}
    }_{{\dot{\bs{\ell}}_i}} = r
    \underbrace{
    \begin{bmatrix}
        \cos(\delta) & -(\theta - \frac{\pi}{2}) \sin(\delta) \\
        -\sin(\delta) & -(\theta - \frac{\pi}{2}) \cos(\delta) \\
        -\cos(\delta) & (\theta - \frac{\pi}{2}) \sin(\delta) \\
        \sin(\delta) & (\theta - \frac{\pi}{2}) \cos(\delta)
\end{bmatrix}
    }_{{\mb{J}_{\ell\psi}}}
    \underbrace{
\begin{bmatrix}
        \dot{\theta} \\
        \dot{\delta}
\end{bmatrix}
    }_{\dot{\bs{\psi}}}
\end{equation}
or, the simplified version,
\begin{equation}
    \dot{\bs{\ell}}_i = r \;{{\mb{J}_{\ell\psi}}}\; \dot{\bs{\psi}}
    \label{act_compensation}
\end{equation}
knowing that, \(\bs{\ell}_i \triangleq R\, \theta_{m,i}\)\,,
\begin{equation}
 \dot{\bs{\ell}}_i = R \; \dot{\theta}_{m,i}
    \label{eq:servo_act_velocity_relation}
\end{equation}

where, \(R\) is the radius of the pulleys as mentioned in Table~\ref{tab:param}, and \(\dot{\theta}_{m,i} \) are rotational velocity of the servo motors and \(r\) is the radius of pitch circle (the radius from the main backbone and the secondary backbone) as shown in Fig.\ref{conintuum_arm}. \\
Comparing \eqref{act_compensation} and \eqref{eq:servo_act_velocity_relation}, the mapping from actuation space to the task space is as follow:
\begin{align}
& \underbrace{
    \begin{bmatrix}
        \dot{\theta}_{m,1} \\
        \vdots \\
        \dot{\theta}_{m,4}
    \end{bmatrix}
    }_{{\dot{\bs{\theta}}_{m,i}}} =
    \underbrace{
    r/R \quad{{\mb{J}_{\ell\psi}}}
    }_{\mb{J}_{a}} \quad 
    \underbrace{
     \begin{bmatrix}
        \dot{\theta} \\
        \dot{\delta}
    \end{bmatrix}
    }_{\dot{\bs{\psi}}} \\
& \dot{\bs{\theta}}_{m,i} = \mb{J}_{a} \dot{\bs{\psi}},  \qquad  \mb{J}_{a} \in \mathbb{R}^{4 \times2} \label{eq:actutaion_jacob}
\end{align} \par
\section{Teleoperation}
\begin{figure*}[!h]
    \centering
    \includegraphics[width=\linewidth]{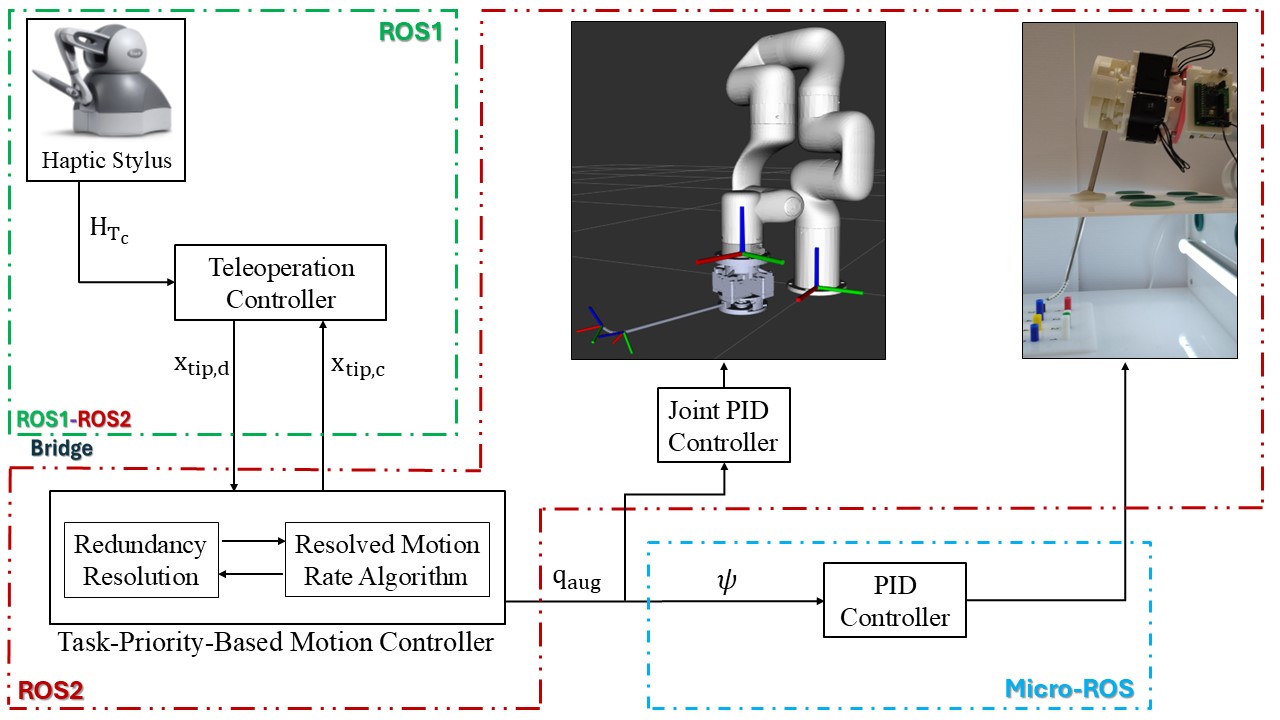}
    \caption{Teleoperation control architecture of the surgical manipulator system}
    \label{fig:control_schematic}
\end{figure*}

This study introduces a leader-follower teleoperation framework developed for user-supervised direct motion control of surgical instruments. The framework, illustrated in Fig.\ref{fig:control_schematic}, enables user interaction through a haptic device, transmitting the stylus’s pose to the teleoperation controller. The controller retrieves the current tip pose from the motion controller in \emph{ROS2} and calculates a relative command pose for the haptic device, based on the reference anchor position and its current location. This calculated pose is then sent as the desired tip position to the leader robot system. \par 
This paper uses a haptic stylus as the leader device for teleoperating the surgical manipulator system. Supporting 6 degrees of freedom, the stylus enables intuitive control of the follower manipulator. Its popularity in haptics and teleoperation research is further enhanced by open-source APIs that facilitate communication with the stylus. The software package referenced in \cite{CRTK2020} includes a \emph{ROS} driver that transmits pose messages from the stylus, along with its two-button states, at a rate of 500 Hz.

\subsection{Haptic Device Trajectory Mapping} \label{haptic}
\begin{figure}[!h]
	\centering
	\includegraphics[width=0.76\textwidth, height=0.34\textheight]{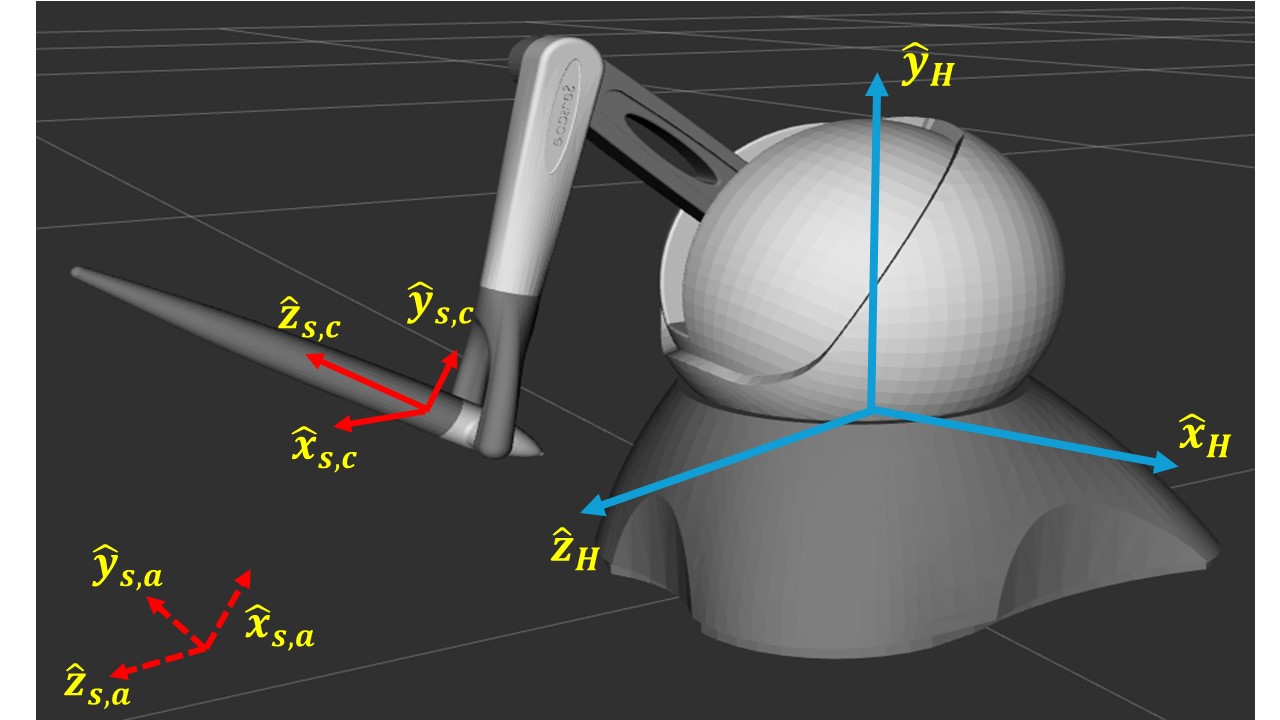}
	\caption{3D Systems Touch haptic stylus reference frames.}
	\label{fig:haptic_frames}
\end{figure}
The proximal buttons on the stylus are configured in various combinations to act as teleoperation command buttons, allowing control of both the manipulator and the continuum segment, either jointly or independently. The pose of the stylus at the start of teleoperation is termed the anchor pose, designated as \(^{\text{H}}\mathbf{T}_{\text{s,a}}\). All subsequent poses when the command button is pressed are referred to as the current pose, represented by \(^{\text{H}}\mathbf{T}_{\text{s,c}}\). A sample visual of these frames is provided in Fig.\ref{fig:haptic_frames}. The relationship between the anchor and current pose of the stylus is expressed as follows:
\begin{equation}
    ^{\text{H}}\mathbf{T}_{\text{s,c}} = {}^{\text{H}}\mathbf{T}_{\text{s,a}}\;{}^{{\text{s,a}}}\mathbf{T}_{\text{s,c/a}}\;
\end{equation}
Here, \(^{\text{s,a}}\mathbf{T}_{\text{s,c/a}}\) represents the change in pose from the anchor frame to the current frame of the stylus, expressed in the anchor frame. In teleoperation, this relative pose change is essential, as it determines the corresponding relative pose change for the surgical manipulator, achieved through a similarity transformation.
\begin{equation}
    ^{\text{tip,a}}\mathbf{T}_{\text{s,c/a}} = ({}^{\text{tip,a}}\mathbf{T}_{\text{s,a}})({}^{\text{s,a}}\mathbf{T}_{\text{s,c/a}}) ({}^{\text{tip,a}}\mathbf{T}_{\text{s,a}})^{-1}\;
\end{equation}

\noindent where \(({\text{tip,a}})\) represents the tip \emph{anchor} frame, corresponding to the surgical instrument's pose at the moment the command button is first engaged. The transformation \({}^{\text{tip,a}}\mathbf{T}_{\text{s,a}}\) is calculated using the following expression:
\begin{equation}
    ^{\text{tip,a}}\mathbf{T}_{\text{s,a}} = ({}^\text{b}\mathbf{T}_{\text{tip,a}})^{-1} ({}^\text{b}\mathbf{T}_{\text{H}}) ({}^{\text{H}}\mathbf{T}_{\text{s,a}})\;
\end{equation}

\noindent where \(^\text{b}\mathbf{T}_{\text{tip,a}}\) represents the anchor pose of the surgical instrument’s \emph{tip}, expressed in the manipulator’s base frame.

This transformed relative pose is crucial for calculating the pose of the tip in the base frame \(\{\text{b}\}\), using the following expression:
\begin{equation}
    ^\text{b}\mathbf{T}_{\text{tip,d}} = {}^\text{b}\mathbf{T}_{\text{tip,a}}\;{}^{\text{tip,a}}\mathbf{T}_{\text{s,c/a}}\; \label{eq:teleop-desired}
\end{equation}

\noindent where \(^\text{b}\mathbf{T}_{\text{tip,d}}\) is a homogeneous transformation matrix representing the desired pose of the tip written in the robot base frame.
\subsection{Resolved Motion Rate Algorithm}
    
The desired pose of the continuum instrument tip is defined as follows:
\begin{equation}
\mathbf{x}_{\text{tip,d}} \triangleq (\mathbf{p}_{\text{tip,d}}, \, \mathbf{R}_{\text{tip,d}})
\end{equation}
Here, \(\mb{p}\) denotes the position of the tip, while \(\mb{R}\) is the \(3\times 3\) rotation matrix representing the tip's orientation. This pose is specified in real time by the operator and controlled using a resolved motion rate algorithm, which computes joint commands to drive the manipulator, as originally introduced in \cite{Whitney1969}.
The position error \( \mb{e}_{\text{p}} \) of the instrument tip is defined as the difference between the current pose \( \mb{x}_{\text{tip,c}} \) and the desired pose \( \mb{x}_{\text{tip,d}} \),  expressed as:
\begin{align}
\mb{e}_\text{p} &= \mb{p}_{\text{tip,d}} - \mb{p}_{\text{tip,c}}
\end{align}
The orientation error \( \mb{e}_{\text{o}} \) of the continuum instrument's tip is determined by first computing the relative error rotation matrix \( \mb{R}_{\text{e}} \), which captures the difference between the current orientation \( \mb{R}_{\text{tip,c}} \) and the desired orientation \( \mb{R}_{{\text{tip,d}}} \), expressed as:

\begin{align}
&\mb{R}_\text{e} = \mb{R}_{\text{tip,d}} \hspace{2pt} {\mb{R}^{\mathrm{T}}_{\text{tip,c}} } \\
&\Theta = \arccos\left(\frac{\text{tr}(\mb{R}_\text{e}) - 1}{2}\right) \\
&\mb{e}_\text{o} = \frac{\Theta}{2 \sin \Theta} 
\begin{bmatrix}
\mb{R}_\text{e}(3,2) - \mb{R}_\text{e}(2,3) \\
\mb{R}_\text{e}(1,3) - \mb{R}_\text{e}(3,1) \\
\mb{R}_\text{e}(2,1) - \mb{R}_\text{e}(1,2)
\end{bmatrix}
\end{align}
\begin{align}
v_{\text{mag}} &= 
\begin{cases} 
    v_{\text{mx}} & \text{if } \|\mb{e}_\text{p}\| > \frac{\epsilon_\text{p}}{\gamma_\text{p}} \\
    v_{\text{mn}} + ( v_{\text{mx}} - v_{\text{mn}}) \chi_\text{v} & \text{if } \|\mb{e}_\text{p}\| \leq \frac{\epsilon_\text{p}}{\gamma_\text{p}}
\end{cases} \\
\omega_{\text{mag}} &= 
\begin{cases} 
    \omega_{\text{mx}} & \text{if } \|\mb{e}_\text{o}\| > \frac{\epsilon_\text{o}}{\gamma_\text{o}} \\
    \omega_{\text{mn}} + (\omega_{\text{mx}} - \omega_{\text{mn}}) \chi_\omega & \text{if } \|\mb{e}_\text{o}\| \leq \frac{\epsilon_\text{o}}{\gamma_\text{o}}
\end{cases}
\end{align}
where, 
\begin{equation}
    \chi_\text{v} = \frac{\|\mb{e}_\text{p}\| - \gamma_\text{p}}{\gamma_\text{p}(\epsilon_\text{p} - 1)}, \qquad
    \chi_\omega = \frac{\|\mb{e}_\text{o}\| - \gamma_\text{o}}{\gamma_\text{o}(\epsilon_\text{o} - 1)}
\end{equation}
Thus, the desired linear and angular twists can be expressed as:
\begin{align}
&\mb{v} = \frac{v_{\text{mag}} \; \mb{e}_\text{p}}{\|\mb{e}_\text{p}\|} \label{eq:v_twist}\\
&\boldsymbol{\omega} = \frac{\omega_{\text{mag}} \; \mb{e}_\text{o}}{\|\mb{e}_\text{o}\|} \label{eq:w_twist}
\end{align}

Resolved motion rate control leverages position and orientation errors to compute the desired instrument tip twist, \( \bs{\xi}_\text{tip} \). As the \textit{tip} nears the target position and the errors reduce to zero, the tip twist correspondingly decreases.\\
In this context, \(v_{\text{mx}}\) and \(v_{\text{mn}}\) represent the maximum and minimum linear velocities of the tip, while \(\omega_{\text{mx}}\) and \(\omega_{\text{mn}}\) denote the maximum and minimum angular velocities, respectively.
\(\gamma_\text{p}\) and \(\gamma_\text{o}\) denote the permissible thresholds for position and orientation errors, while \(\epsilon_\text{p}\) and \(\epsilon_\text{o}\) represent the relative error bounds within which the tip shifts from a resolved rate to a reduced proportional rate.\par
\begin{figure*}[!t]
    \centering
    \renewcommand{\thesubfigure}{\alph{subfigure}}
    \begin{subfigure}[b]{1\columnwidth}
        \centering
        \includegraphics[width=\linewidth, height=0.38\textheight]{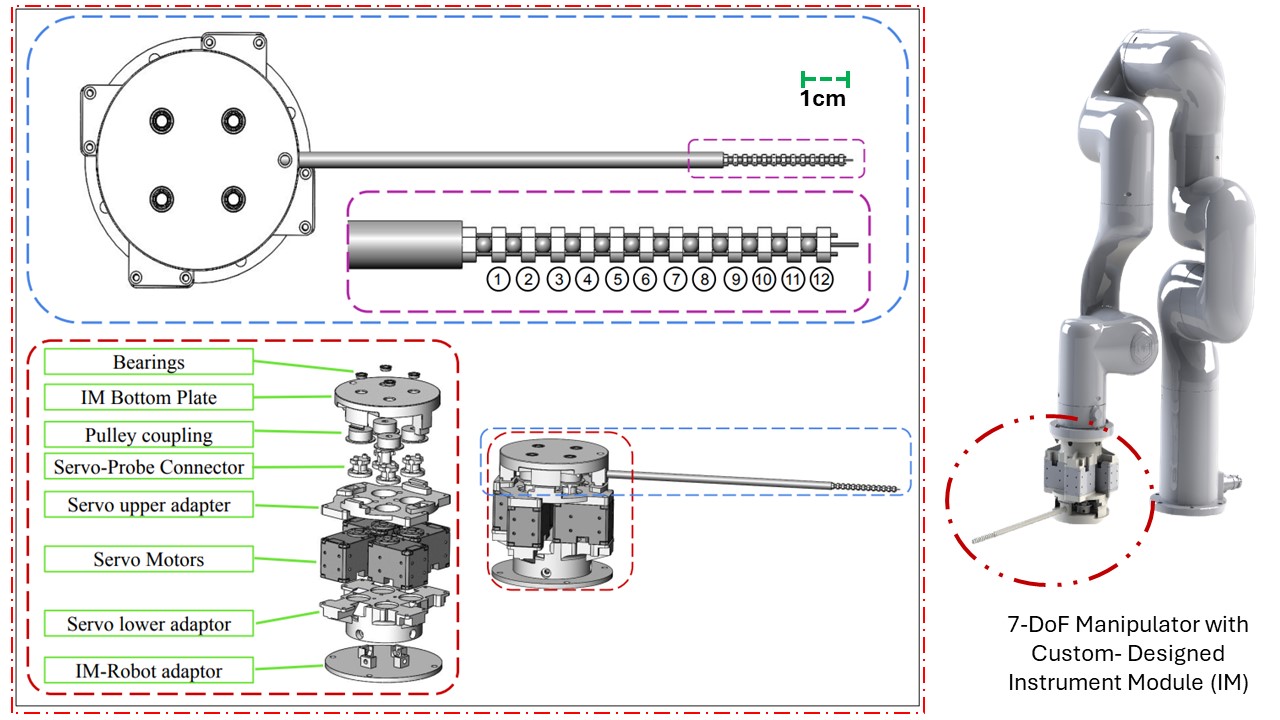}
        \caption{The custom-designed instrument module (IM) features a cable-driven surgical continuum instrument: the assembled CAD model of the 7-DoF manipulator with the IM; top view of the module; and the exploded view.}
        \label{fig:Multiple_views}
    \end{subfigure}
    \vfill
    \begin{subfigure}[b]{1\columnwidth}
        \centering
        \includegraphics[width=\linewidth, height=0.38\textheight]{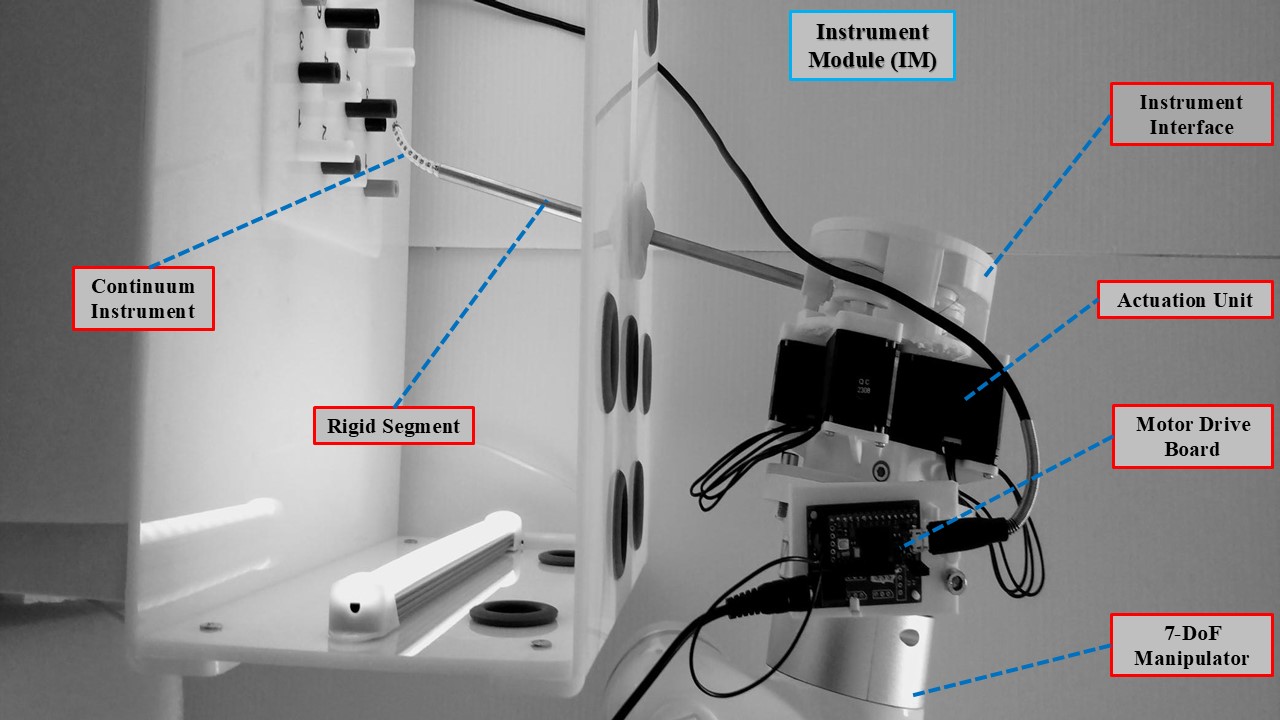}
        \caption{Instrument module architecture design includes a custom-designed surgical instrument, an actuation unit, an instrument interface, and a motor drive board.}
        \label{fig:IDM_arch}
    \end{subfigure}
    \caption{CAD and 3D-printed models of the custom-designed instrument module (IM)}
    \label{fig:combined}
\end{figure*}
\subsection{System Control Architecture}
Figure\ref{fig:control_schematic} illustrates the control architecture used for teleoperating the surgical manipulator system. The haptic stylus transmits device state information to the teleoperation controller operating in a ROS1 environment. The motion controller, running in ROS2, incorporates resolved motion rate control with a redundancy resolution method. Communication between the teleoperation controller and motion controller is managed through a ROS1-ROS2 bridge. The instrument module (IM) is actuated by servos managed through micro-ROS, which extends ROS2 capabilities to the microcontroller. This control system is engineered for precise and responsive teleoperation, and its modular architecture, integrating both ROS1 and ROS2, allows us to utilize existing open-source software packages for real-time control.
\section{Hardware Development of an Instrument Module} \label{IDM}
The \emph{instrument module} (IM) functions to enable articulation control of the surgical continuum instrument, as illustrated in Fig.\ref{fig:combined}. This section details the mechatronic development of the instrument module.
\subsection{Design Features}

The proposed low-cost, open-source instrument module for operating a custom-designed, cable-driven miniature continuum instrument incorporates the following design features:
\begin{itemize}
     \item \textit{Compactness}: The module is lightweight and compact, making it suitable for mounting on and manipulation by a collaborative robot.
     
    \item \textit{Actuation Redundancy}: The 2-DoF of the 3cm continuum instrument is effectively managed by four NiTi cables actuated by four servo motors. This actuation redundancy reduces the peak tendon pulling load on each individual servo. 
    
    \item \textit{ROS Compatibility}: The module is built within the micro-ROS framework, minimizing the delay between operator commands and real-time actuation of the surgical instrument during teleoperation.
\end{itemize} \par
\subsection{Mechanical Design}
The actuation unit controls the surgical instrument through a four-pulley interface. The proposed module achieves two degrees of freedom actuation using four Dynamixel XL-430 servo motors. While the minimum number of pulling cables to actuate a continuum instrument is two, we use four cables, and the actuation redundancy provides a larger wrench-feasible workspace in configuration space and generally reduces the peak tendon pulling load per individual. Figure\ref{fig:Multiple_views} presents the exploded view of the CAD assembly, including the IM and the custom-designed surgical instrument. The design specifications of the proposed IM are detailed in Table.\ref{tab:param}. The hardware prototype of the IM is shown in Fig.\ref{fig:Multiple_views}, with the CAD files of the design made available as open source \footnote{\url{https://github.com/stevens-armlab/Teleop_Continuum_IM}}.
\begin{table}[ht]
    \centering
    \captionsetup{justification=centering}
    \caption{Instrument module specifications}
    \renewcommand{\arraystretch}{1.1} %
    \setlength{\tabcolsep}{6pt} %
    \begin{tabular}{ p{0.44\columnwidth} | p{0.44\columnwidth} }
        \hline
        \textbf{Component} &  \textbf{Specification} \\
        \hline
        Servo model & Dynamixel XL-430-W250 \\[0.5ex]
        Motor adapter dimension & 89.7 $\times$ 85.6 (mm) \\[0.5ex]
        Number of links & 12 \\[0.5ex]
        Number of cables & 4 \\[0.5ex]
        Diameter of disks & 4.60 (mm) \\[0.5ex]
        Length of continuum segment & 30 (mm) \\[0.5ex]
        Length of rigid segment & 230 (mm) \\[0.5ex]
        Diameter of the motor pulley & 10.8 (mm) \\[0.5ex]
        Diameter of Nitinol backbones & 0.23 (mm) \\[0.5ex]
        \hline
    \end{tabular}
    \label{tab:param}
\end{table}
\subsection{Electronics Development}
As described in Section \ref{haptic} and depicted in Fig.\ref{fig:control_schematic},  the haptic device enables user commands for the instrument tip held by the robot manipulator. The main control node and the resolved rate algorithm operate in ROS2, while the haptic device utilizes ROS1 to transmit the desired instrument pose to the ROS2 node through a ROS1-ROS2 bridge. A Teensy board, integrated with Micro-ROS, facilitates communication with the ROS2 node to control the servos in the surgical instrument's actuation unit.
Micro-ROS extends the robot operating system framework to microcontrollers, enabling the Teensy board to interface with ROS2. This integration facilitates servo control, thereby enabling precise manipulation of the surgical instrument.\par
We employed the \textit{Teensy} board as the control board for the actuation unit. A custom Dynamixel shield was integrated with the Teensy board, utilizing UART serial communication to ensure reliable and high-speed servo control. Furthermore, we configured \textit{Teensyduino}, an Arduino software add-on, to program and operate the Teensy microcontrollers within the Arduino IDE.
\begin{table}[h]
\caption{Overview of Experimental Validation for Priority Cases}
\centering
\begin{tabularx}{\textwidth}{|X|c|c|c|}
\hline
\textbf{Aspect} & \textbf{Case0} & \textbf{Case1} & \textbf{Case2} \\
\hline
Trajectory Tracking Experiment & $\checkmark$ & $\checkmark$ & $\checkmark$ \\
\hline
Ring Board Experiment & $\checkmark$ & $\checkmark$ & $\checkmark$ \\
\hline
Silicon Board Experiment & & $\checkmark$ & $\checkmark$ \\
 \hline
 Actuation Unit Experiment & $\checkmark$ & & \\
\hline
\end{tabularx}
\label{table:EXP_cases}
\end{table}
\begin{figure*}
    \centering
    \includegraphics[width=\linewidth]{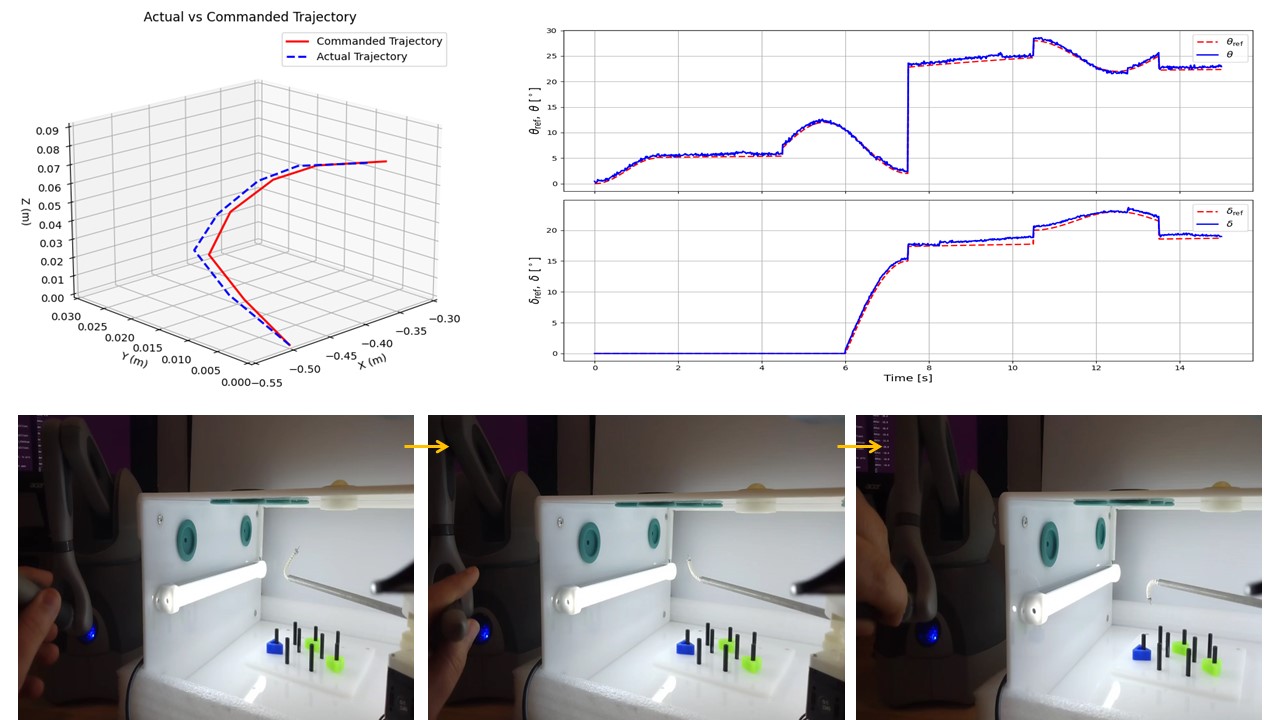}
    \caption{The actual versus commanded trajectory of the \emph{tip}, with the robot remaining locked, along with the continuum instrument's \(\theta\) and \(\delta\) angles (actual versus reference values)}
    \label{fig:Case4_error}
\end{figure*}
\begin{figure}[!h]
    \centering
    \renewcommand{\thesubfigure}{\alph{subfigure}}  
    \begin{subfigure}[b]{1\columnwidth}
        \centering
        \includegraphics[width=\linewidth, height=0.36\textheight]{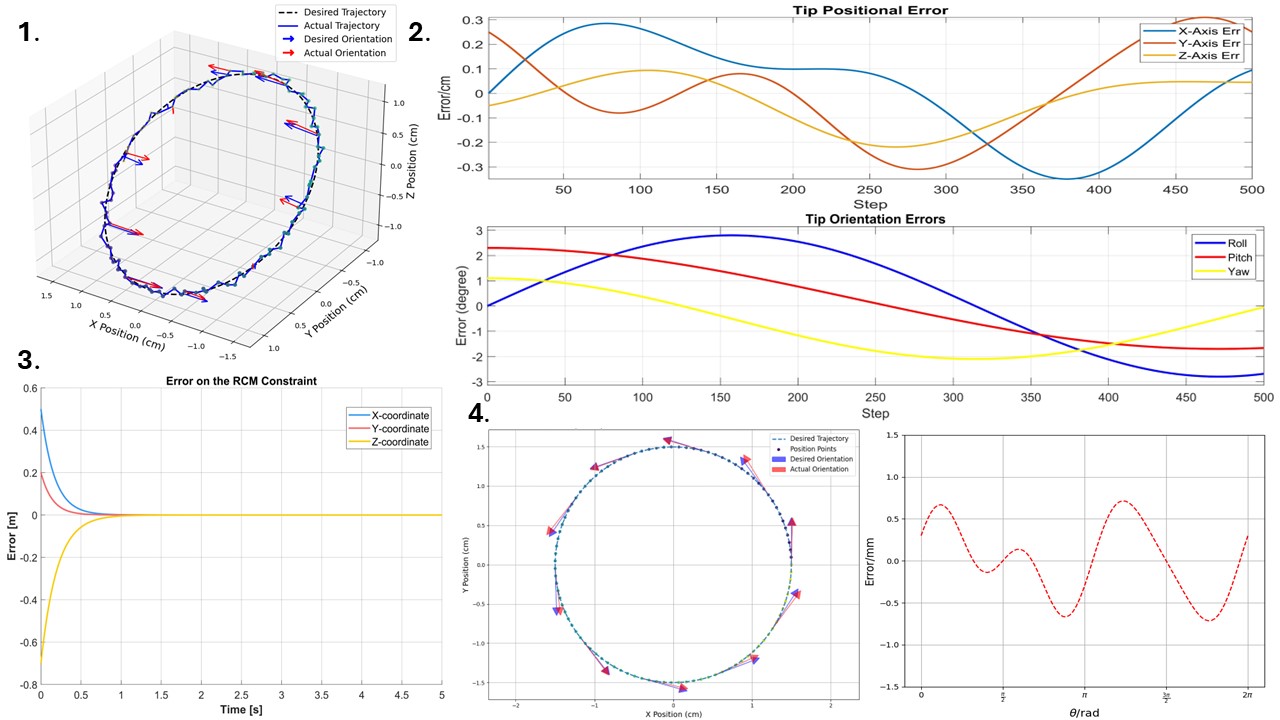}
        \caption{Circular-shaped path trajectory in simulation: 1) The circular path trajectory(actual vs. desired pose), 2) The position and orientation errors, 3) The RCM constraint error and, 4) A 2nd circular path trajectory in y-z plane and its error.}
        \label{fig:traject_sim_circle}
    \end{subfigure}
    \hfill
    \begin{subfigure}[b]{\columnwidth}
        \centering
        \includegraphics[width=\linewidth,,height=0.36\textheight]{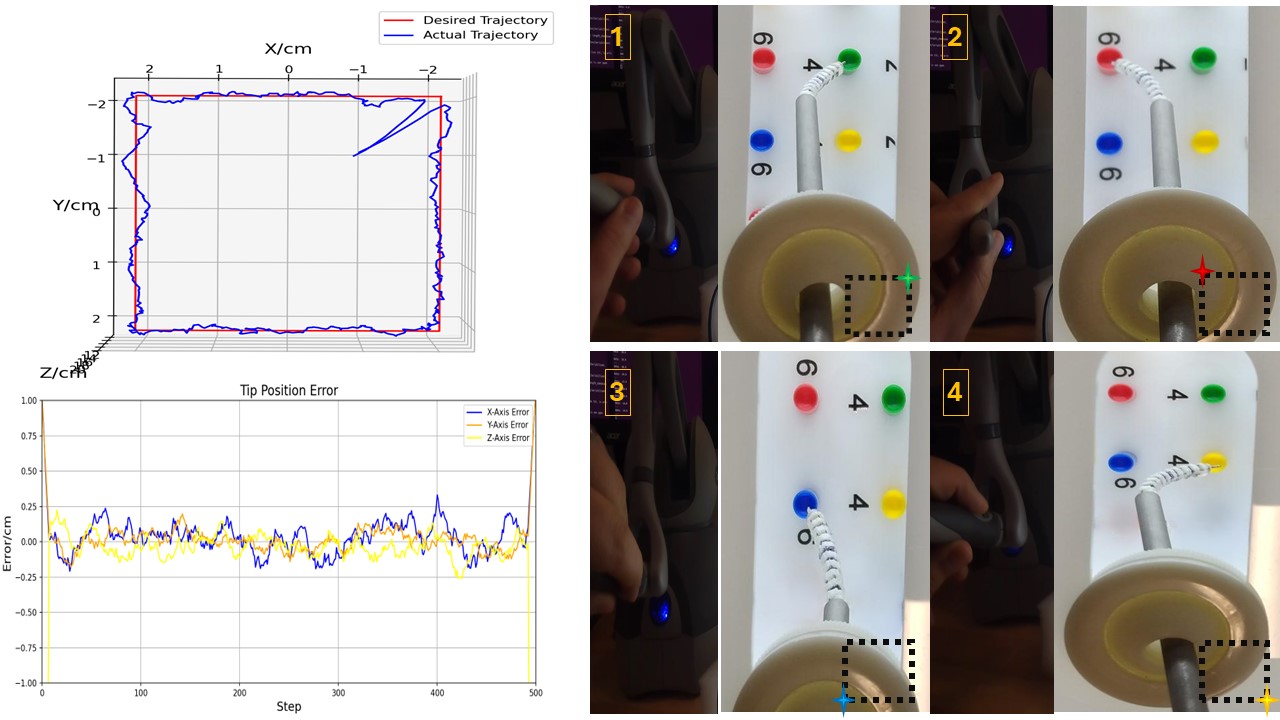}
        \caption{The positional error of the \textit{tip} in the hardware experiment for a square-shaped path tracking utilizing a pegboard.}
        \label{fig:traject_sim_sqaure}
    \end{subfigure}
    \caption{Trajectory tracking simulation and experiment for the proposed kinematic framework}
    \label{fig:traject_sim}
\end{figure}
\section{Experimental Results}
\begin{figure*}
    \centering
    \includegraphics[width=1.1\linewidth,height=0.5\textheight, trim=80 0 70 0, clip]{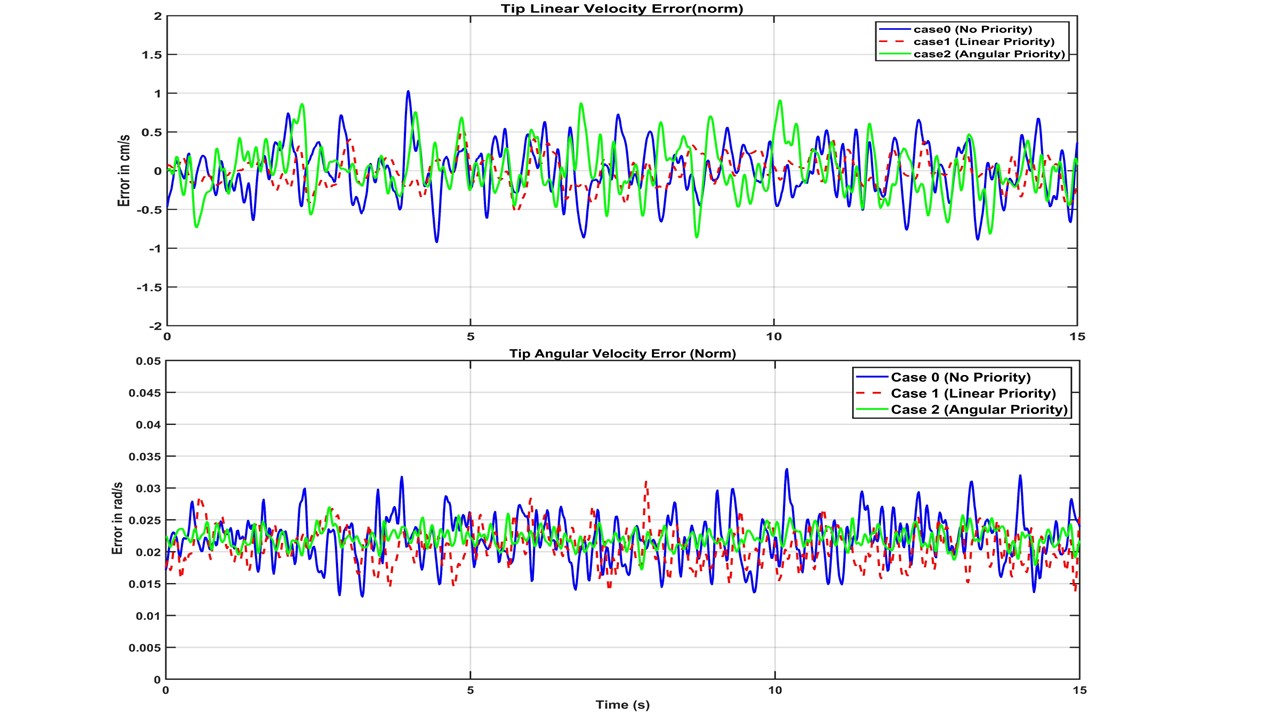}
    \caption{Error in task priority cases for linear and angular velocity teleoperation commands}
    \label{fig:priority_error_plots}
\end{figure*}
\subsection{Experimental Setup and Overview}
The experimental setup, shown in Fig.\ref{fig:System_arch}, consists of a 7-DoF manipulator, a custom-designed instrument module with a 3cm miniature continuum instrument of 4.6mm diameter, a motor driving board that controls the actuation unit, a haptic Touch stylus device, and a surgical kit. We have investigated the teleoperation of the continuum instrument kinematics by exploring the task priorities while the main task is to satisfying the RCM constraint. For this purpose, simulations were conducted in ROS2 environment. Additionally, hardware experiments were carried out to validate the framework. Table.\ref{table:EXP_cases} provides an overview of the experiments and the corresponding cases used in each experiment. \par
\subsection{Preliminary Instrument Module Actuation Unit Test}
At the beginning, for the experimental evaluations, we considered an arc-shaped path of 5cm in total for the tip to follow. we aimed to lock the robot's motion, without assigning priority, and focused solely on the accuracy of the actuation unit during teleoperation. As shown in Fig.\ref{fig:Case4_error} we attempted to teleoperate the continuum segment along the path, while the robot remained locked. Referring to \eqref{eq:tip_velocity}, the propose framework using \eqref{eq:actutaion_jacob}, provides a good accuracy in teleoperating the servos to control the continuum segment. During the first 6 seconds, we attempted to bend the tip (\(\theta\)), We then commanded the tip's tilt angle \(\delta\), to follow the commanded trajectory with high accuracy using \eqref{eq:actutaion_jacob}. The maximum error for bending angles was 6 degrees for \(\delta\) and approximately 3 degrees for \(\theta\).\par

\subsection{Trajectory Tracking Experiment for Case0 }
To assess the computational efficiency and motion accuracy of the developed framework, a few trajectories was defined within the workspace, including a circular path, a square path among others.
For instance, as is shown in Fig.\ref{fig:traject_sim}, we using \eqref{eq:redundancy_res} to control the robot and its continuum segment to follow a circular trajectory while respecting the RCM constraint. This is demonstrated using case0, where no priority is assigned to linear or angular commands. In Fig.\ref{fig:traject_sim_circle}, The robot is teleoperated to follow a circular path with a 3cm diameter. The simulation results show a maximum positional error (norm) of 2.8 mm and an orientation error of 0.05 radian. The error reduced to less than 1mm, in a circular path trajectory in y-z plane as can be seen in Fig.\ref{fig:traject_sim_circle}-4.\\
For the second trajectory test in Fig.\ref{fig:traject_sim_sqaure}, we performed hardware validation as well, this time using a square-shaped path for the \textit{tip} to follow from one peg to another. A pegboard with colorful pegs (green, red, blue, and yellow) was used, with dimensions of 2.3cm between consequent pegs. The fixed path between each consecutive peg was segmented into 500 points, with the direction vector for each point being parallel to the square plane's normal in the z-direction, allowing for the position information of each point along the path to be obtained. The maximum RMSE position error along the path was found to be 3.2mm.
\subsection{Experiment on The Kinematic Performance of Continuum Instrument with Task Priority  for Cases 1 and 2}
\begin{figure}
    \centering
    \includegraphics[width=1.1\columnwidth, height=0.6\columnwidth]{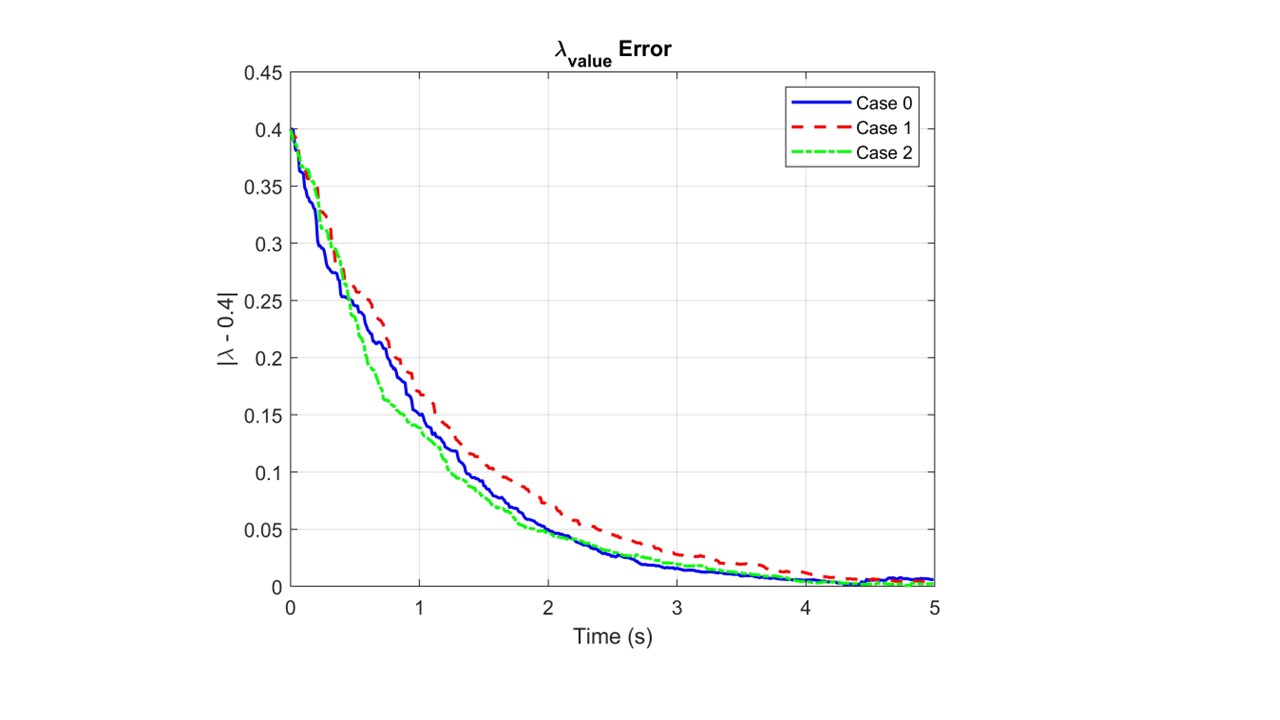}
    \vspace{-35pt} 
    \caption{Comparison of lambda value errors for tasks with and without priority, considering the desired \(\lambda_0 = 0.4\)}
    \label{fig:lambda_variation}
\end{figure}

In order to experimentally evaluate the teleoperated kinematic performance using the task priority cases in section~\ref{task_priroity}, we conducted experiments based on these cases and compared the results with the tasks that had no priorities. Figure.\ref{fig:priority_error_plots} shows a comparison between case0 (no priority) and cases1 and 2 in following the circular trajectory described in Fig.\ref{fig:traject_sim_circle}. As can be seen, the teleoperation linear twist command in case1 has the lowest error, while for the angular twist command, as expected, case2 has the lowest error since it was prioritized over the linear command. Additionally, both prioritized cases have lower twist command errors compared to case0, where no prioritization was applied. For the level-1 task (satisfying the RCM constraint), Fig.\ref{fig:lambda_variation} shows that all three cases behave almost similarly in maintaining \(\lambda\), as close as possible to the desired reference value \(\lambda_0\), as mentioned in \eqref{eq:objective}.\par 

\begin{figure}
    \centering
    \includegraphics[width=\columnwidth]{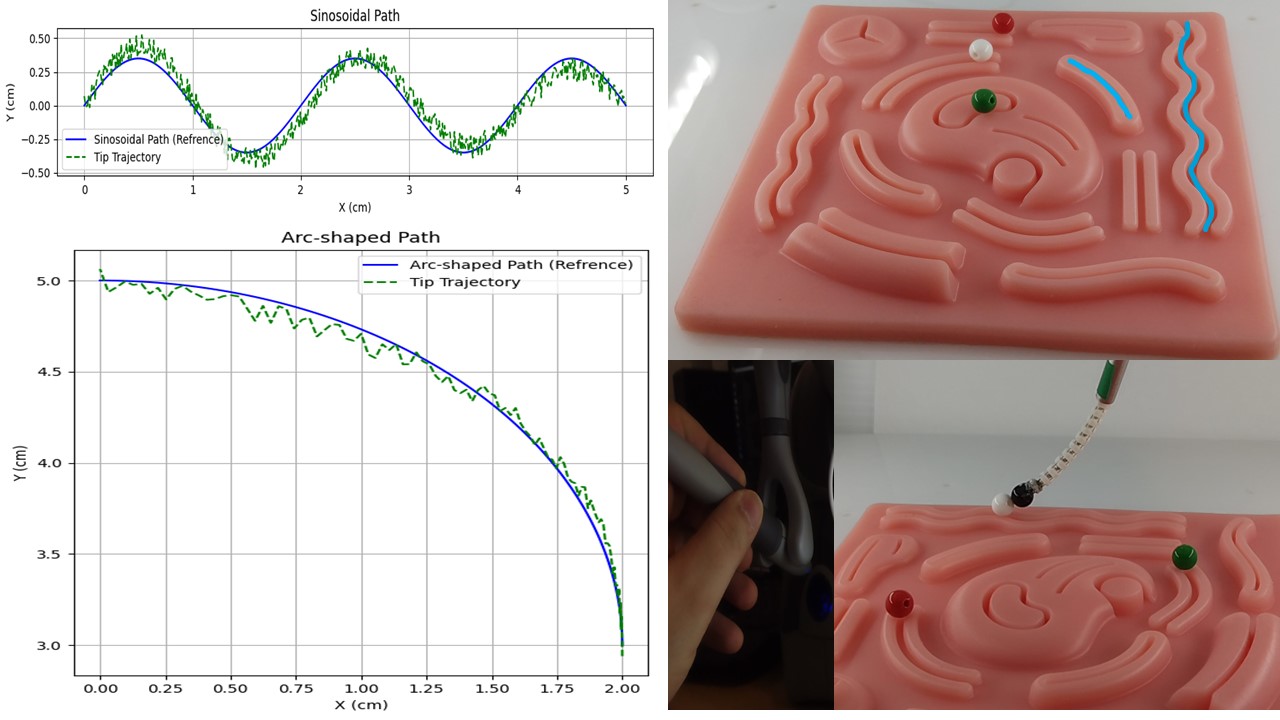}
    \caption{Tip trajectories: the blue solid line represents the reference values, while the green dashed lines indicate the values in hardware experiment.}
    \label{fig:paths_comand_comparison}
\end{figure}

Another test we conducted involved a \emph{silicon board} with carved paths,  illustrated in Fig.\ref{fig:paths_comand_comparison}. In this experiment, we used equation \eqref{eq:final_task_pririty_case1} to guide the tip in pushing a 5mm-radius ball along two different paths: a sinusoidal path  with a 5cm length and 0.3cm amplitude and a arc-shape path with a 2cm radius. As seen from the results, the tip follows the desired user command with high accuracy. The maximum position and orientations errors (norm) was 3.2mm and 6 degree respectively, which occurred on the arc path. The tip trajectory was straightforward for the sinusoidal path due to its small amplitude. However, for the arc-shaped path, the tip orientation needed to change along the trajectory, as this experiment was conducted under case1, where the linear twist of the tip had a higher priority than the angular twist. For comparison, we repeated the experiment under case2, where angular twist was given higher priority than linear twist. The orientation error was reduced to 5 degrees, and the maximum position error remained at approximately 5mm. Supporting videos of the demonstrations are available on YouTube~\footnote{\url{https://youtu.be/m0iEg4JY2vc} , \hspace{1pt} \url{https://youtu.be/vpvyUXYlibA}}.\par
\begin{figure}
    \centering
    \includegraphics[width=\columnwidth,height=0.42\textheight]{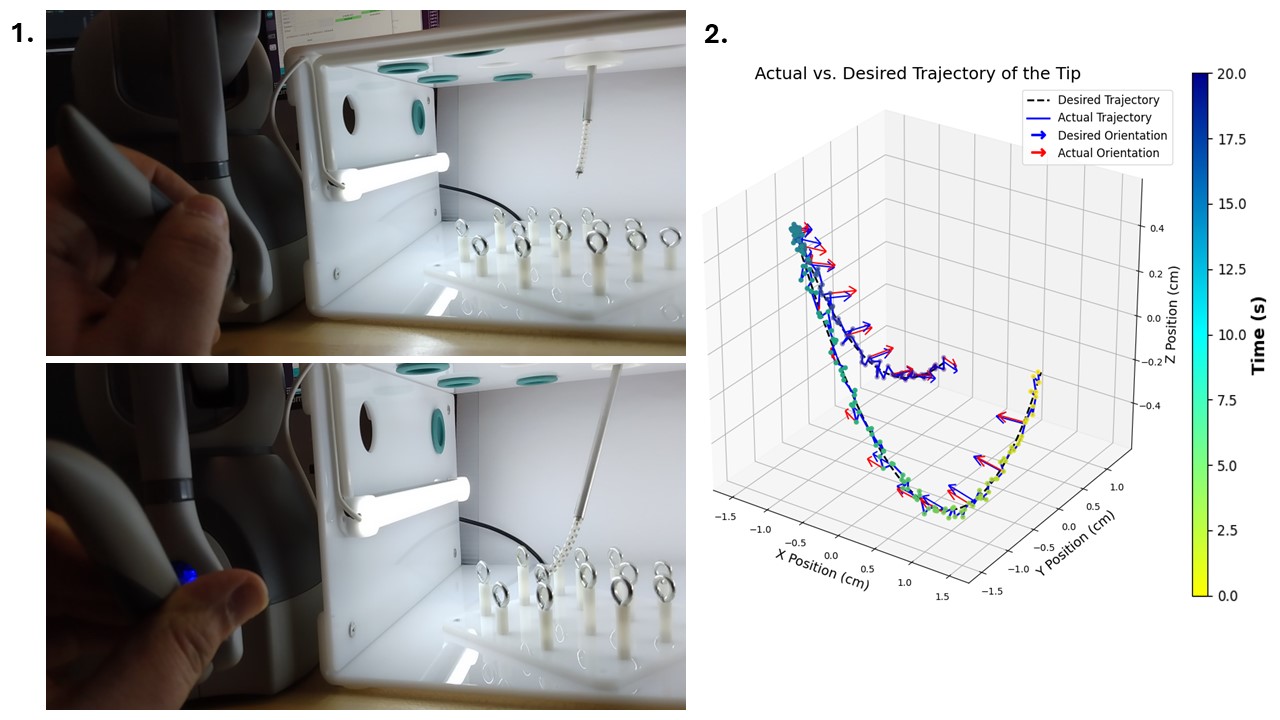}
    \caption{Comparison of the tip's actual and commanded poses using a ring board: (1) Ring board setup, and (2) Tip actual vs. commanded trajectory.}
    \label{fig:ring_board_experiment}
\end{figure}

The third experiment was conducted on a \textit{ring board}, with each ring having a diameter of 8.1mm. The aim was to guide the continuum instrument \emph{tip}, with a diameter of 4.6mm, halfway through the rings for all priority cases. This experiment required significant angular and linear command control. Figure.\ref{fig:ring_board_experiment} shows the results for case2, As observed, the system demonstrated excellent performance, achieving a tip maximum position error of 0.6cm and an orientation error of 2.45 degrees. For case1, the errors were 0.3cm for position and 5.04 degrees for orientation. 
Referring to the methods and metrics explained in Section.\ref{appdx2}, an overview of the performance analysis of the presented framework in these cases is provided in Table.\ref{table:dex}. This overview outlines the use of priority cases in the ring board setup experiment. 
Using the manipulability and inverse condition number metrics, we evaluate \textit{case1}, which prioritizes linear commands, achieving up to 13.5\% and 11.02\% improvements in inverse condition number and manipulability, respectively. For \textit{case2}, we observe approximately an 11\% improvement in numerical stability and a 7\% advantage in manipulability. In both \textit{case1} and \textit{case2}, there is a minimum 5\% improvement in the accuracy of both linear and angular task execution.
\begin{table*}
\centering
\caption{Tip performance comparison for cases 0, 1, and 2 in the ring board experiment } 
\resizebox{\textwidth}{!}{
\begin{tabular}{p{3cm}p{4.5cm}p{4.5cm}p{4.5cm}}
\toprule
\textbf{Aspect} & \textbf{Case 0} & \textbf{Case 1} & \textbf{Case 2} \\
\midrule

\textbf{Task Prioritization} &

\textbf{1. RCM constraint}

\textbf{2. Combined linear-angular twist}

 \(\mathbf{J} \dot{\mathbf{q}}_{\text{aug}} \triangleq \begin{bmatrix} \mathbf{v} \\ \boldsymbol{\omega} \end{bmatrix}\)
&

\textbf{1. RCM constraint}

\textbf{2. Linear twist}

\; \; \(\mathbf{J}_L \dot{\mathbf{q}}_{\text{aug}} = \mathbf{v}\)

\textbf{3. Angular twist}

\; \; \(\mathbf{J}_A \dot{\mathbf{q}}_{\text{aug}} = \boldsymbol{\omega}\)
&
\textbf{1. RCM constraint}

\textbf{2. Angular twist}

\; \; \(\mathbf{J}_A \dot{\mathbf{q}}_{\text{aug}} = \boldsymbol{\omega}\)

\textbf{3. Linear twist}

\; \; \(\mathbf{J}_L \dot{\mathbf{q}}_{\text{aug}} = \mathbf{v}\)
\\
\midrule

\textbf{Inv. Cond. Number, \(\kappa^{-1}\)} &

\textbf{Intermediate Value}

\(\kappa^{-1}_L = 0.60\)

\(\kappa^{-1}_A = 0.58\)

Balanced numerical stability for "both tasks"
&
\textbf{Slightly Higher for Linear}

\(\kappa^{-1}_L = 0.71\)

\(\kappa^{-1}_A = 0.62\)

13.5\% Better numerical stability for "linear" task
&
\textbf{Slightly Higher for Angular}

\(\kappa^{-1}_L = 0.61\)

\(\kappa^{-1}_A = 0.68\)

10.8\% Better numerical stability for "angular" task
\\
\midrule
\textbf{Manipulability (\(m\))} &

\textbf{Balanced Manipulability}

\(m_{\text{L}} = 0.632\)

\(m_{\text{A}} = 0.621\)
&
\textbf{Higher \(m\) for Linear twist}

\(m_{\text{L}} = 0.698\)

\(m_{\text{A}} = 0.625\)
&
\textbf{Higher \(m\) for Angular twist}

\(m_{\text{L}} = 0.635\)

\(m_{\text{A}} = 0.682\)
\\
\midrule

\textbf{Implications} &

\textbf{Uniform Performance}

Executes linear and angular tasks almost equally well
&
\textbf{Best for Linear Precision}

Enhanced better stability and dexterity for linear tasks
&
\textbf{Best for Angular Precision}

Enhanced better stability and dexterity for angular tasks
\\
\midrule

\textbf{Overall Difference} &

\textbf{Intermediate between Cases 1 and 2}

offer a compromise between the linear and angular tasks
&
\textbf{ Better for Linear Tasks}

11\% advantage in executing linear task
&
\textbf{ Better for Angular Tasks}

7\% advantage in executing angular task
\\
\bottomrule
\end{tabular}
}\label{table:dex}
\end{table*}

\section{Conclusion}
This paper tackles the challenge of teleoperating continuum instruments for minimally invasive surgery (MIS). A novel task-priority-based kinematic formulation is developed and employed to quantitatively evaluate teleoperation commands for continuum instruments operating under remote center of motion (RCM) constraints. By leveraging a redundancy resolution approach, the kinematic performance during teleoperation is analyzed, focusing on the comparison of linear and angular commands within a task-priority framework.
For this purpose, we developed a teleoperation framework along with a novel compact, low-cost and modular instrument module (IM), consisting of an actuation unit, motor interface, motor driving board, and a custom-designed surgical instrument. The instrument is composed of 20cm rigid segment and a 3cm miniature continuum instrument. 
The simulations, and experimental validations were conducted using a standard surgical kit to demonstrate the effectiveness of the developed framework.\par
Using redundancy resolution methods, we formulate three levels of task-priority, enabling the prioritization of linear versus angular commands at the continuum instrument \textit{tip} in diverse scenarios.
The initial experiment tested the teleoperation framework by commanding the actuation unit to guide the tip along an arc-shaped path while keeping the manipulator stationary. This setup was designed to evaluate the teleoperation performance of the actuation unit. Subsequent experiments focused on path tracking using the proposed framework without task prioritization. Both simulation and hardware implementations yielded promising results, with the tip successfully following the desired commands via teleoperated twists generated using the haptic Touch device. The maximum positional error along the path was 2.8 mm for the tip position, and the maximum orientation error was 0.05 radians. The same task was then repeated under different priority cases to facilitate comparative analysis.

Additional experiments were conducted using both case1 and case2 on a silicon board with predefined patterns. In these tests, teleoperated commands successfully guided the tip to push small, colorful balls (4 mm radius) along the prescribed paths. Furthermore, a ring board was used, where the tip was commanded to bend precisely, enabling it to pass successfully and accurately through multiple rings with an 8.1 mm diameter, arranged at various angles.

Finally, a performance analysis was conducted, comparing all cases by decomposing the Jacobian of the continuum instrument tip using Singular Value Decomposition (SVD). The singular values were used to compute two metrics: the \emph{inverse condition number} and \emph{manipulability}. These metrics were employed to evaluate the dexterity and performance of the proposed task-priority motion control framework. The results indicated that prioritizing linear commands improved accuracy and performance by over 11\%, while prioritizing angular commands provided a 7\% advantage in executing angular tasks compared to linear ones.
\section{Limitation and Future Work}
In this study, we present a quantitative investigation of teleoperation commands for continuum instruments as an open research challenge in robot-assisted MIS—one that, to the best of our knowledge, is being explored for the first time in the literature and has not yet been comprehensively addressed. To achieve this, we utilized the task-priority method within redundancy resolution and formulated a unique 3-level task-priority approach specifically designed to support this investigation. Our approach also enables experimental validation as an initial step toward addressing this open discussion in robot-assisted MIS.
However, this study is limited in that it does not include clinical tests. The experiments were conducted in controlled, simulated environments, which may not fully capture the complexity and variability of real-world clinical scenarios. This underscores aspects of our current research that should be further explored in future clinical studies. 
We plan to conduct further clinical studies, including cadaver and potentially animal studies, to validate our findings in more realistic surgical scenarios. In addition, future research will also focus on further decoupling specific elements within the linear and angular command sets while simultaneously prioritizing them to enhance dexterity in targeted directions. This approach is particularly crucial for improving the operation of continuum instruments in confined surgical environments.

\section*{Acknowledgment}
This research was supported in part by NSF Grant CMMI-2138896. 

We would like to thank Guoqing Zhang for his contribution to the design of the Instrument Module in this project.
\section{Appendix 1 }  \label{appdx1}
In this appendix, we formulate a two-level task-priority problem. For example, the task-priority redundancy resolution method considers the robot's instantaneous kinematic equations as the primary objective, which must be satisfied before addressing secondary tasks. %
\begin{align}
\textbf{Task level 1:} \quad & \mathbf{J}_1 \dot{\mathbf{q}} = \dot{\mathbf{x}}_1 \label{eq:task1} \\
\textbf{Task level 2:} \quad & {\text{minimize}} \; \|\mathbf{J}_2 \dot{\mathbf{q}} - \dot{\mathbf{x}}_2 \|
\end{align}

For example, $\dot{\mb{x}}_1$ may be the desired end-effector twist and $\dot{\mb{x}}_2$ may be $\nabla \mb{g}$ (a desired value of the objective function derivative, typically zero).

Solving task 1, we get:
\begin{equation}
\dot{\mb{q}} = \mb{J}_1^\dagger \dot{\mb{x}}_1 + \left(\mb{I} - \mb{J}_1^\dagger \mb{J}_1\right) \bs{\eta}_1
\label{eq:solution_task1}
\end{equation}

Substituting $\dot{\mb{q}}$ into task 2, we solve for $\bs{\eta}_1$ to minimize the error in task 2:
\begin{equation}
\mb{J}_2 \left(\mb{J}_1^\dagger \dot{\mb{x}}_1 + \left(\mb{I} - \mb{J}_1^\dagger \mb{J}_1\right)\bs{\eta}_1\right) = \dot{\mb{x}}_2
\end{equation}

Solving for $\bs{\eta}_1$, we get:
\begin{equation}
\bs{\eta}_1 = \Tilde{\mb{J}}_1^\dagger \left(\dot{\mb{x}}_2 - \mb{J}_2 \mb{J}_1^\dagger \dot{\mb{x}}_1 \right) + \left(\mb{I} - \Tilde{\mb{J}}_1^\dagger \Tilde{\mb{J}}_1 \right)\bs{\eta}_2
\label{eq:solution_eta1}
\end{equation}

where $\Tilde{\mb{J}}_1^\dagger = \mb{J}_2 \left(\mb{I} - \mb{J}_1^\dagger \mb{J}_1\right)$. If we want to optimize a third task, we can choose $\bs{\eta}_2$ in a similar way as we did for $\bs{\eta}_1$. If $\mb{I} - \Tilde{\mb{J}}_1^\dagger \Tilde{\mb{J}}_1 = 0$, then we have used up all the available redundancy in the robot.

Substituting the particular solution of $\bs{\eta}_1$ into \eqref{eq:solution_task1}, we get:
\begin{equation}
\begin{split}
\dot{\mathbf{q}} =\, & \mathbf{J}_1^\dagger \dot{\mathbf{x}}_1 + \left(\mathbf{I} - \mathbf{J}_1^\dagger \mathbf{J}_1\right)  \left(\mathbf{J}_2 \left(\mathbf{I} - \mathbf{J}_1^\dagger \mathbf{J}_1\right)\right)^\mathrm{T} \left(\dot{\mathbf{x}}_2 - \mathbf{J}_2 \mathbf{J}_1^\dagger \dot{\mathbf{x}}_1 \right) \\
& \quad + \left(\mathbf{I} - \tilde{\mathbf{J}}_1^\dagger \tilde{\mathbf{J}}_1 \right)\boldsymbol{\eta}_2 
\end{split}
\end{equation}
We use the idempotence property, $(\mb{I} - \mb{J}_1^\dagger \mb{J}_1)^\dagger = \left(\mb{I} - \mb{J}_1^\dagger \mb{J}_1\right)$, and we also use $(\mb{I} - \mb{J}_1^\dagger \mb{J}_1)^2 = \mb{I} - \mb{J}_1^\dagger \mb{J}_1$, property. After these simplifications, we obtain:
\begin{equation}
\dot{\mb{q}} = \mb{J}_1^\dagger \dot{\mb{x}}_1 + \left(\mb{I} - \mb{J}_1 \mb{J}_1^\dagger\right)\mb{J}_2^\dagger \; ( \dot{\mb{x}}_2 - \mb{J}_2 \mb{J}_1^\dagger \dot{\mb{x}}_1)
\label{eq:simplified_solution}
\end{equation}

The general solution, including the homogeneous term from equation above, becomes:
\begin{equation}
\dot{\mb{q}} = \mb{J}_1^\dagger \dot{\mb{x}}_1 +\Tilde{\mb{J}}_1^\dagger \left(\dot{\mb{x}}_2 - \mb{J}_2 \mb{J}_1^\dagger \dot{\mb{x}}_1\right) + \left(\mb{I} - \mb{J}_1^\dagger \mb{J}_1\right)\left(\mb{I} - \Tilde{\mb{J}}_1^\dagger \Tilde{\mb{J}}_1\right) \bs{\eta}_2
\label{eq:general_solution}
\end{equation}
where $\bs{\eta}_2$ may be determined to satisfy a tertiary task. 

The solution in equation \eqref{eq:general_solution} simplifies, using the pseudo-inverse properties, to the following result:
\begin{equation}
\dot{\mb{q}} = \mb{J}_1^\dagger \dot{\mb{x}}_1 + \Tilde{\mb{J}}_1^\dagger \left(\dot{\mb{x}}_2 - \mb{J}_2 \mb{J}_1^\dagger \dot{\mb{x}}_1\right) + \left(\mb{I} - \mb{J}_1^\dagger \mb{J}_1 - \Tilde{\mb{J}}_1^\dagger \Tilde{\mb{J}}_1^\dagger \right) \bs{\eta}_2
\label{eq:simplified_final_solution}
\end{equation}
\section{Appendix 2} \label{appdx2}
In this appendix, we report on the performance of the proposed framework.
 In robotics, a system's dexterity significantly affects its applicability. Within this framework, the manipulability measure is a widely recognized index used to assess the system's ability to maneuver within the workspace. This measure is associated with the concept of manipulability ellipsoids, first introduced in \cite{yoshikawa1985}, where the volume of the ellipsoid serves as an indicator of the uniformity in the mapping between the joint space and task space. This ellipsoid are used to analyze the kinematic feasibility of generating arbitrary end-effector velocities or forces for a given joint configuration. This measure is expressed as:
\begin{equation}
    m= \sqrt{\text{det}(\mb{J} \; \mb{J}^{\mathrm{T}})} \label{eq:manipulability}
\end{equation}
Referring to the task-priority cases in \eqref{eq:final_task_pririty_case1} and \eqref{eq:final_task_pririty_case2}, we consider this index analysis using the \(6\times 10\) Jacobian of \emph{tip}, 
\(\mb{J'}_\mathrm{T}\),examining its linear and angular components separately.\par
In order to report the teleoperation motion commands dexterity, we can decompose the Jacobian \(\mb{J'}_\mathrm{T}\) using singular value decomposition (SVD), to explore both the joint and task subspaces. This decomposition is defined as follows:
  \begin{equation}
      \mb{J} \triangleq \mb{U} \bs{\Sigma} \mb{V}^{\mathrm{T}}
      \label{eq:decompose_jacob}
  \end{equation}
 where,

 \begin{itemize}
     \item $\mb{U}$ (an $m \times m$ matrix) contains the left singular vectors and corresponds to "motion in task space".
     \item $\bs{\Sigma}$ (an $m \times n$ diagonal matrix) contains the singular values.
     \item $\mb{V}^{\mathrm{T}}$ (an $n \times n$ matrix) contains the right singular vectors and corresponds to "motion in joint space".
 \end{itemize} \par
The term \( \mb{J} \mb{J}^{\mathrm{T}}\) in \eqref{eq:manipulability} is a symmetric matrix whose determinant equals the square of the product of the singular values of \( \mb{J} \). This is because the singular values of \( \mb{J} \), denoted as \( s_i \), are related to the eigenvalues \( G_i \) of \( \mb{J} \mb{J}^{\mathrm{T}} \) by \( G_i = s_i^2 \). Therefore, \( \text{det}(\mb{J}  \mb{J}^{\mathrm{T}}) \triangleq \prod s_i \), which aligns with the product of singular values. \\
The geometric interpretation of \( \sqrt{\text{det}(\mb{J} \mb{J}^{\mathrm{T}})} \) is the volume of the manipulability ellipsoid. This volume is an indicator of how well the robot can manipulate its environment in all directions emanating from a given configuration. A larger volume indicates greater manipulability, implying the robot can more effectively move in various directions without encountering singular configurations. It should be noted that the ratio of the minimum singular value to the maximum singular value, defined as the inverse condition number, represents the radius of the manipulability ellipsoid geometrically and serves as a valuable indicator of a robot manipulator's dexterity, precision, and stability.
\begin{equation}
    \kappa^{-1} = s_{\text{min}} / s_{\text{max}}
\end{equation}
A performance overview of the proposed framework in this study for the \emph{ring board} experiment of Fig.\ref{fig:ring_board_experiment} is presented in Table.\ref{table:dex}.

\bibliographystyle{IEEEtran}
\bibliography{ref}

\end{document}